\RequirePackage{fix-cm}
\documentclass[twocolumn,natbib]{svjour3}          
\smartqed  
\usepackage{graphicx}
\usepackage{lineno, blindtext}
\usepackage{times}
\usepackage{epsfig}
\usepackage{amsmath}
\usepackage{amssymb}
\usepackage{cuted}
\usepackage{capt-of}
\usepackage{textcomp}
\usepackage{multirow}
\usepackage[ruled]{algorithm2e}

\usepackage[breaklinks=true,bookmarks=false]{hyperref}

\newcommand{\norm}[1]{\left\lVert#1\right\rVert}

\DeclareMathOperator*{\argmin}{\arg\!\min}
 
\begin{document}
\begin{sloppypar}

\title{3DFaceGAN: Adversarial Nets for 3D Face Representation, Generation, and Translation
}

\author{Stylianos Moschoglou\textsuperscript{*} \and Stylianos Ploumpis\textsuperscript{*} \and Mihalis Nicolaou \and Athanasios Papaioannou \and Stefanos Zafeiriou} 

\authorrunning{S. Moschoglou et al.}

\institute{\textsuperscript{*} The authors contributed equally to this work.\\ \\
S. Moschoglou, S. Ploumpis, A. Papaioannou and S. Zafeiriou are with the Department of Computing, Imperial College London, United Kingdom and Facesoft.io.\\
E-mail: \{s.moschoglou, s.ploumpis, a.papaioannou11, s.zafeiriou\}@imperial.ac.uk\\
M. Nicolaou is with the Computation-based Science and Technology Research Centre, The Cyprus Institute.\\
E-mail: m.nicolaou@cyi.ac.cy
}

\date{Received: date / Accepted: date}

\maketitle


\begin{figure*}[t]
\begin{center}
\includegraphics[width=0.9\linewidth]{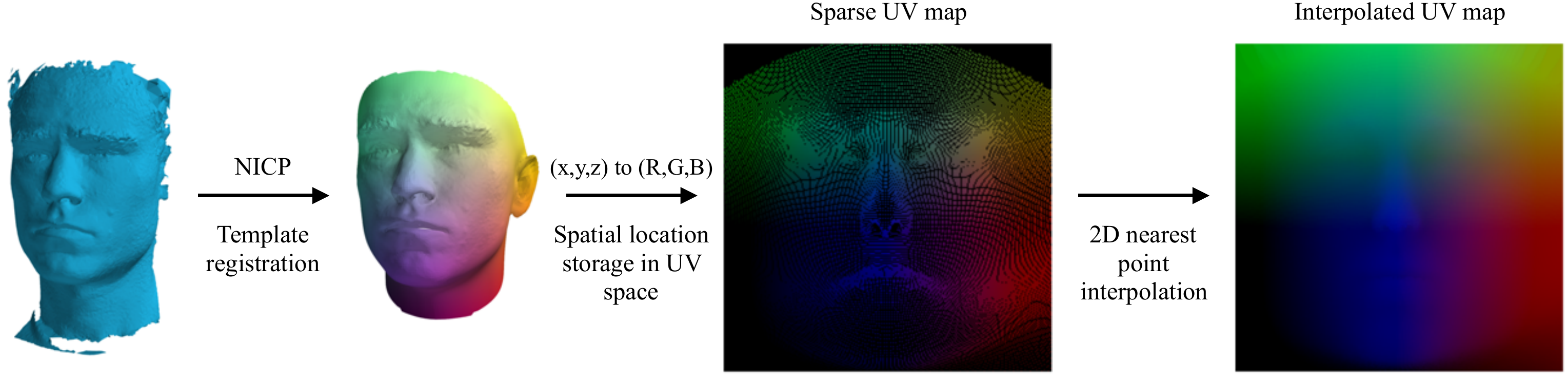}
\caption{A graphical representation of the data preprocessing step. We begin by applying non-rigidly a mesh template to the raw scan and we later store the spatial information of the vertices $(x,y,z)$ into a UV space. Lastly, a 2D nearest point interpolation is performed to fill out the missing values.}
\label{fig:_preprocessing_uvs}
\end{center}
\end{figure*}

\begin{abstract}
Over the past few years, Generative Adversarial Networks (GANs) have garnered increased interest among researchers in Computer Vision, with applications including, but not limited to, image generation, translation, imputation, and super-resolution. Nevertheless, no GAN-based method has been proposed in the literature that can successfully represent, generate or translate 3D facial shapes (meshes). This can be primarily attributed to two facts, namely that (a) publicly available 3D face databases are scarce as well as limited in terms of sample size and variability (e.g., few subjects, little diversity in race and gender), and (b) mesh convolutions for deep networks present several challenges that are not entirely tackled in the literature, leading to operator approximations and model instability, often failing to preserve high-frequency components of the distribution. As a result, linear methods such as Principal Component Analysis (PCA) have been mainly utilized towards 3D shape analysis, despite being unable to capture non-linearities and high frequency details of the 3D face - such as eyelid and lip variations. In this work, we present 3DFaceGAN, the first GAN tailored towards modeling the distribution of 3D facial surfaces, while retaining the high frequency details of 3D face shapes. We conduct an extensive series of both qualitative and quantitative experiments, where  the merits of 3DFaceGAN are clearly demonstrated against other, state-of-the-art methods in tasks such as 3D shape representation, generation, and translation.
\end{abstract}

\section{Introduction}

GANs are a promising unsupervised machine learning methodology implemented by a system of two deep neural networks competing against each other in a zero-sum game framework \Citep{goodfellow2014generative}. GANs became immediately very popular due to their unprecedented capability in terms of implicitly modeling the distribution of visual data, thus being able to generate and synthesize novel yet realistic images and videos, by preserving high-frequency details of the data distribution and hence appearing authentic to human observers. Many different GAN architectures have been proposed over the past few years, such as the Deep Convolutional GAN (DCGAN) \Citep{radford2015unsupervised} and the Progressive GAN (PGAN) \Citep{karras2017progressive}, which was the first to show impressive results in generation of high-resolution images. 

A type of GANs which has also been extensively studied in the literature is the so-called Conditional GAN (CGAN) \Citep{mirza2014conditional}, where the inputs of the generator as well as the discriminator are conditioned on the class labels. Applications of CGANs include domain transfer \Citep{kim2017learning, bousmalis2017unsupervised, tzeng2017adversarial}, image completion \Citep{li2017generative, yang2017high, wang2017shape}, image super-resolution  \Citep{nguyen2018super, johnson2016perceptual, ledig2017photo} and image translation \Citep{isola2017image, zhu2017unpaired, choi2017stargan, wang2018high}. 

Despite the great success GANs have had in 2D image/video generation, representation, and translation, no GAN method tailored towards tackling the aforementioned tasks in 3D shapes has been introduced in the literature. This is primarily attributed to the lack of appropriate decoder networks for meshes that are able to retain the high frequency details  \Citep{dosovitskiy2016generating, jackson2017large}.

\begin{figure*}[t]
\includegraphics[width=1\textwidth]{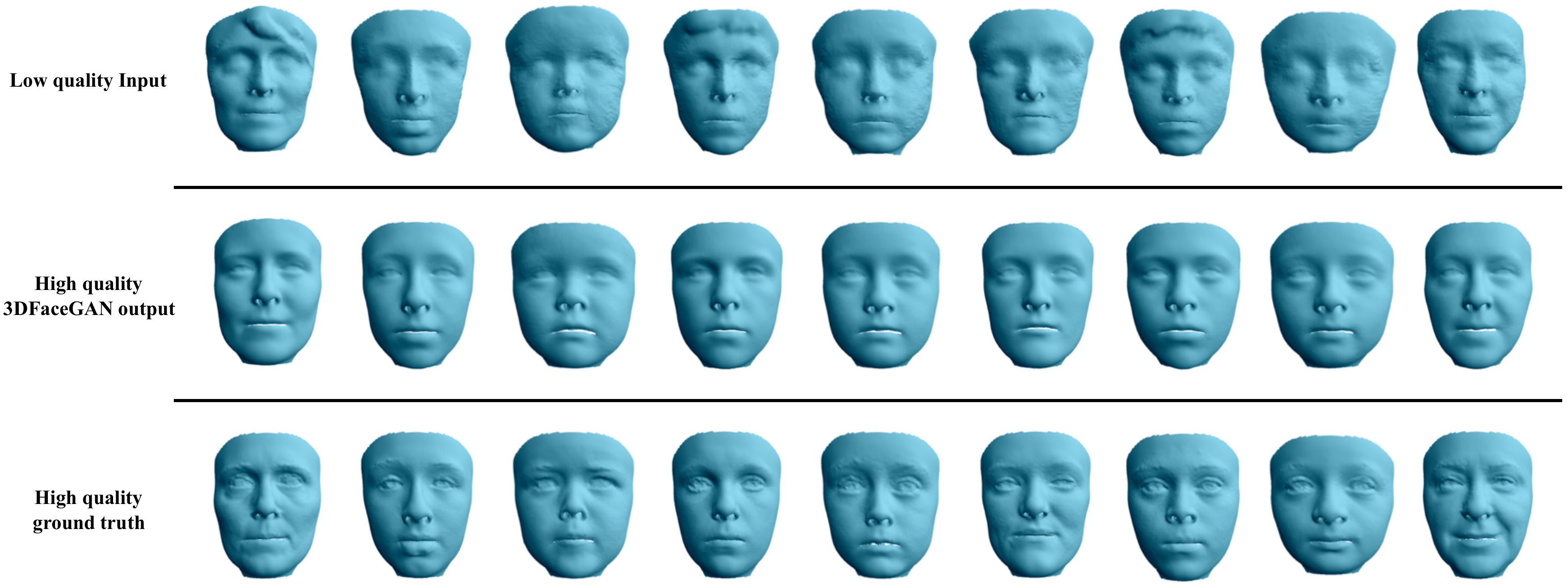}
\captionof{figure}{\small{Results of 3DFaceGAN in the shape translation task on test data of the proposed Hi-Lo database. The first row of shapes shows the low quality facial meshes captured by a low cost sensor, whereas the bottom row depicts the same subjects captured in high quality by an expensive high-end apparatus. The middle row shows our shape translation output results when the network takes as inputs the low quality 3D facial scans.}
\label{fig:low_high}}
\end{figure*}

\begin{figure*}[t]\centering
\includegraphics[width=0.9\textwidth]{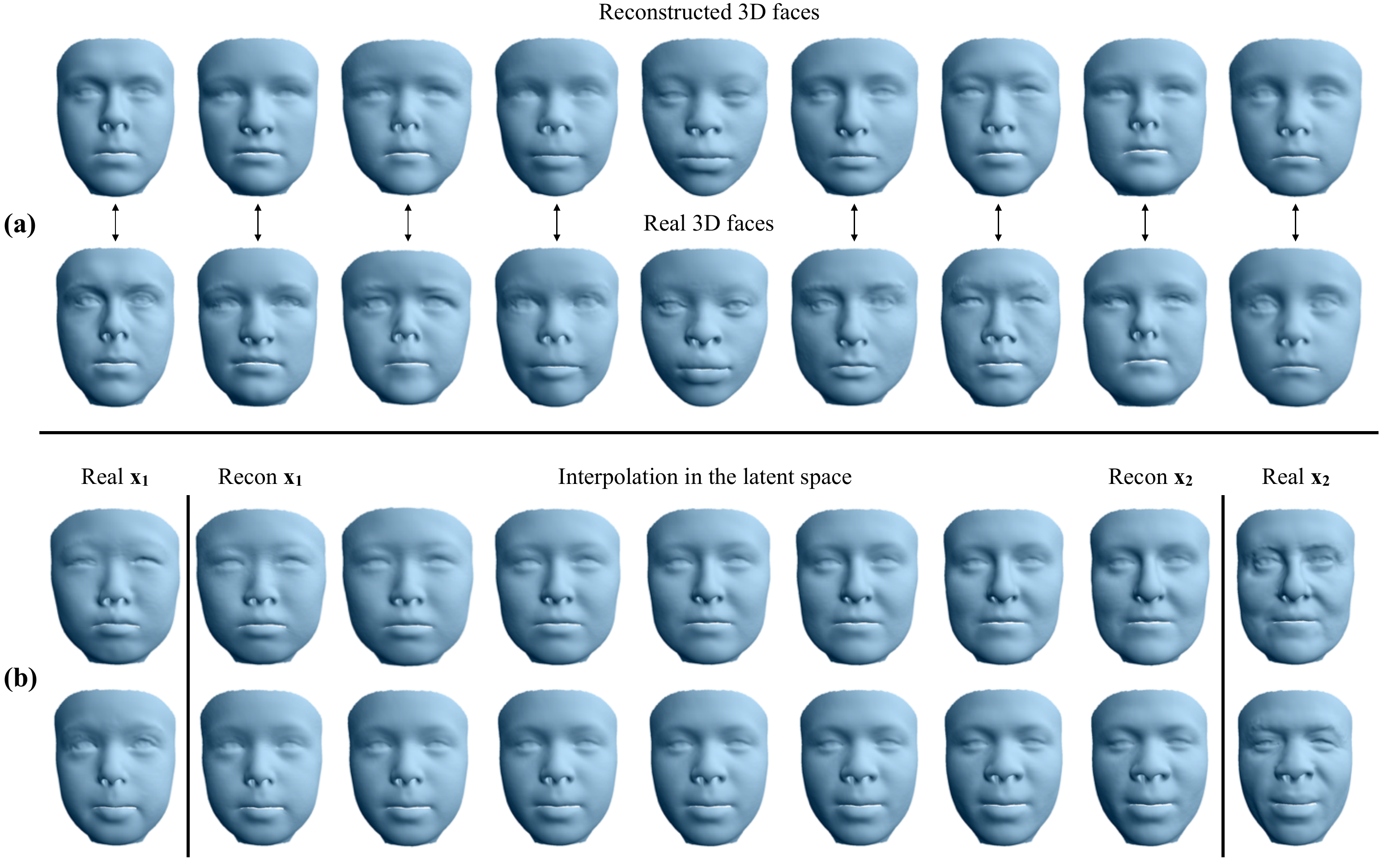}
\captionof{figure}{3D face representation and generation utilizing the proposed 3DFaceGAN. In (a) we demonstrate the 3D face representation capability of 3DFaceGAN. The first row shows the reconstructed 3D faces whereas the second row shows the corresponding real 3D faces. As evidenced, 3DFaceGAN is able to capture and reconstruct non-linear details of the 3D face such as lips, eyelids, etc. In (b) we present the generative nature of 3DFaceGAN. The left and right hand side show the real 3D face targets. The generated samples in between show the reconstructions and the interpolations of the targets in the latent space.}
\label{fig:rep_inter}
\end{figure*}

In this paper, we study the task of representation, generation, and translation of 3D facial surfaces using GANs. Examples of the applications of 3DFaceGAN in the tasks of 3D face translation as well as 3D face representation and generation are presented in Fig. \ref{fig:low_high} and Fig. \ref{fig:rep_inter}, respectively. Due to the fact that (a) the use of volumetric representation leads to very low-quality representation of faces \Citep{fan2017point, qi2017pointnet}, and (b) the current geometric deep learning approaches \Citep{bronstein2017geometric}, and especially spectral convolution, preserve only the low-frequency details of the 3D faces, we study approaches that use 2D convolutions in a UV unwrapping of the 3D face. The process of unwrapping a 3D face in the UV domain is shown in Fig. \ref{fig:_preprocessing_uvs}. Overall, the contributions of this work can be summarized as follows.

\begin{itemize}
    \item We introduce a novel autoencoder-like network architecture for GANs, which achieves state-of-the-art results in tasks such as 3D face representation, generation, and translation.
    \item We introduce a novel training framework for GANs, especially tailored for 3D facial data.
    \item We introduce a novel process for generating realistic 3D facial data, retaining the high frequency details of the 3D face.
\end{itemize}

The rest of the paper is structured as follows. In Section \ref{sec:3D_rep}, we succinctly present the various methodologies that can be utilized in order to feed 3D facial data into a deep network and argue why the UV unwrapping of the 3D face was the method of choice. In Section \ref{sec:3DFaceGAN}, we present all the details with respect to 3DFaceGAN training process, losses, and model architectures. Finally, in Section \ref{sec:experiments}, we provide information about the database we collected, the preprocessing we carried out in the databases we utilized for the experiments and lastly we present extensive quantitative and qualitative experiments of 3DFaceGAN against other state-of-the-art deep networks.

\section{3D face representations for deep nets}
\label{sec:3D_rep}

The most natural representation of a 3D face is through a 3D mesh. Adopting a 3D mesh representation requires application of mesh convolutions defined on non-Euclidean domains (i.e., geometric deep learning methodologies\footnote{A thorough overview describing the first attempts towards geometric deep learning can be found in \cite{bronstein2017geometric}.}). Over the past few years, the field of geometric deep learning has received significant attention \Citep{maron2017convolutional,litany2017deep,lei2017deriving}. Methods relevant to this paper are auto-encoder structures such as \cite{ranjan2018generating,litany2017deformable}. Nevertheless, such auto-encoders, due to the type of convolutions applied, mainly preserve low-frequency details of the meshes. Furthermore, architectures that could potentially preserve high-frequency details, such as skip connections, have not yet been attempted in geometric deep learning. Therefore, geometric deep learning methods are not yet suitable for the problem we study in this paper.  

\begin{figure*}[t]
\begin{center}
\includegraphics[width=0.9\linewidth]{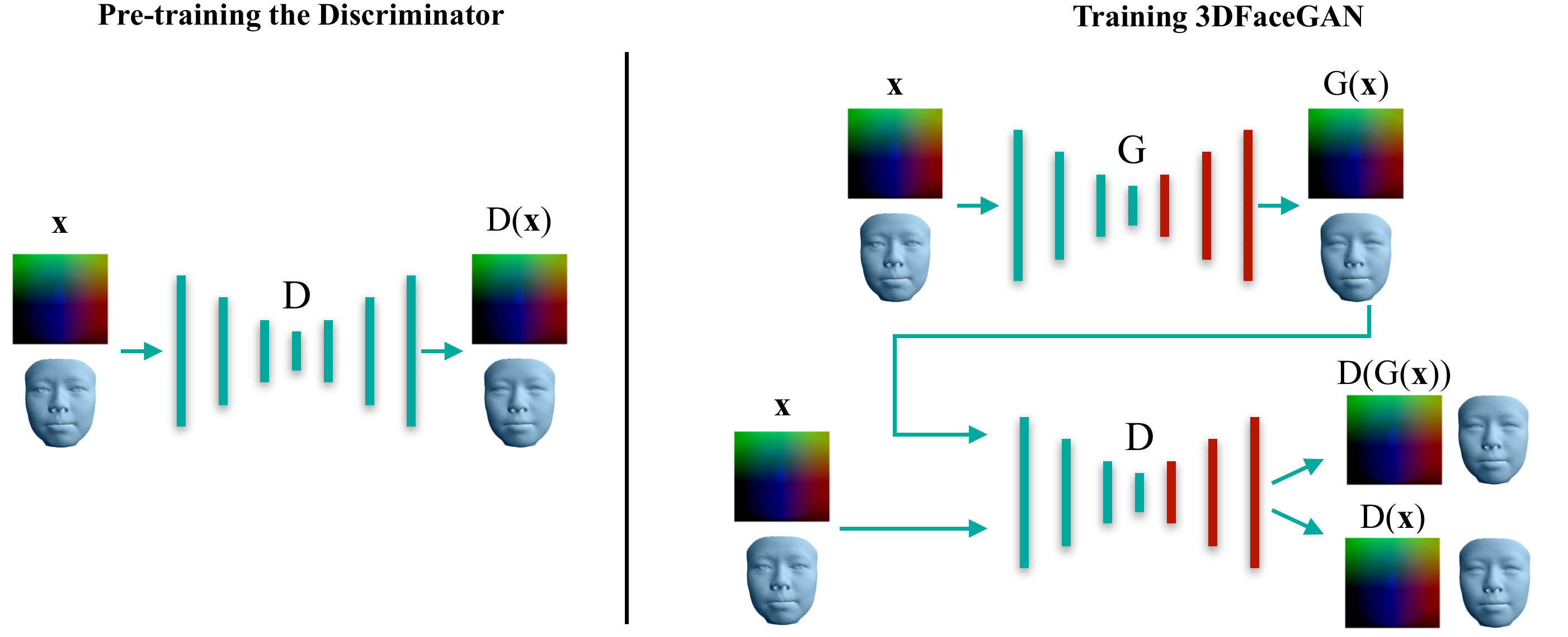}
\caption{3DFaceGAN training process in a nutshell. The networks receive (extract) 2D facial UVs as inputs (outputs). The corresponding 3D faces are shown below or next to them. We firstly pre-train $D$ (left figure). We then use the learned weights/biases to initialize $D$ and $G$ and subsequently start the adversarial training (right figure). The decoder parts of $D$ and $G$ are depicted in red color as we freeze the weights/biases updates during the training phase of 3DFaceGAN.}
\label{fig:3DFaceGAN_overview}
\end{center}
\end{figure*}

Another way to work with 3D meshes is to concatenate the coordinates of the 3D points in an 1D vector and utilize fully connected layers to decode correctly the structure of the point cloud \Citep{fan2017point, qi2017pointnet}. Nevertheless, in this way the triangulation and spatial adjacent information is lost and the number of the parameters describing this formulation is extremely large which makes the network hard to train. 

Recently, many approaches aim at regressing directly on the latent parameters of a learned model space, e.g., PCA, rather than the 3D coordinates of points \Citep{richardson2017learning,tran2017regressing,dou2017end,genova2018unsupervised}. This formulation limits the geometrical details of the 3D representations and is restricted to their latent model space. In contrast, a 3D volumetric space is introduced in \cite{jackson2017large} as a representation of a 3D structure and exploits a Volumetric Regression Network which outputs a discretized version of the 3D structure. Due to discretization, the predicted 3D shape has low quality and corresponds to non-surface points that are difficult to handle.

Lastly, in \cite{feng2018joint}, a UV spatial map framework is utilized where the 3D coordinates of the points are stored in a UV space instead of the texture values of the mesh. This formulation exhibits a very good representation for 3D meshes where there are no overlapping regions and the mesh is optimally unwrapped. Since the 3D mesh is transferred in a 2D UV domain, we are then able to use 2D convolutions, with the whole range of capabilities they offer. As a result, this is our preferred methodology for preprocessing the 3D face scans, as further explained in Section \ref{sec:data_preprocessing}. 

\section{3DFaceGAN}
\label{sec:3DFaceGAN}
In this Section we describe the training process, network architectures, and loss functions we utilized for 3DFaceGAN. Moreover, we discuss the framework we utilized for 3D face generation as well as present an extension of 3DFaceGAN which is able to handle data annotated with multiple labels.

\subsection{Objective function}
The main objective of the generator $G$ is to retrieve a facial UV map $x$ as input and generate a \textsl{fake} one, $G\left(x\right)$, which in turn should be as close as possible to the \textsl{real} target facial UV map $y$. For example, in the case of 3D face translation, the input can be a neutral face and the output a certain expression (e.g., \textsl{happiness}) or in the case of 3D face reconstruction the input can be a 3D facial UV map and the output a reconstruction of the particular 3D facial UV map. The goal of the discriminator $D$ is to distinguish between the \textsl{real} ($y$) and \textsl{fake} ($G\left(x\right)$) facial UV maps. Throughout the training process, $D$ and $G$ compete against each other until they reach an equilibrium, i.e., until $D$ can no longer differentiate between the \textsl{fake} and the \textsl{real} facial UV maps.

\begin{changemargin}{0cm}{0cm} 
\textbf{Adversarial loss.} To achieve the 3DFaceGAN objective, we propose to utilize the following loss for the adversarial part. That is,
\begin{align}
      \begin{array}{ll}
        \mathcal{L}_D = \mathbb{E}_{y}\left[\mathcal{L}\left(y\right)\right] - \lambda_{adv}\cdot\mathbb{E}_{x}\left[\mathcal{L}\left(G(x)\right)\right],\label{eq:adv_loss}\\
          \mathcal{L}_G = \mathbb{E}_{x}\left[\mathcal{L}\left(G(x)\right)\right],
        \end{array}
\end{align}
where $D\left(\cdot\right)$ refers to the output of the discriminator $D$, $\mathcal{L}\left(x\right)\doteq \norm{x - D(x)}_1$, and $\lambda_{adv}$ is the hyper-parameter which controls how much weight should be put on $\mathcal{L}\left(G(x)\right)$. The higher the $\lambda_{adv}$, the more emphasis $D$ puts on the task of differentiating between the real and fake data. The lower the $\lambda_{adv}$, the more emphasis $D$ puts on reconstructing the actual real data. There is a fine line between which task $D$ should primarily focus on by adjusting $\lambda_{adv}$. In our experiments we deduced that for relatively low values of $\lambda_{adv}$ we retrieve optimal performance as then $D$ is able to influence the updates of $G$ in such a way that the generated facial UV maps are more realistic. During the adversarial training, $D$ tries to minimize $\mathcal{L}_{D}$ whereas $G$ tries to minimize $\mathcal{L}_G$. Similar to recent works such as \cite{zhao2016energy, berthelot2017began}, the discriminator $D$ has the structure of an autoencoder. Nevertheless, the main differences are that (a) we do not make use of the margin $m$ as in \cite{zhao2016energy} or the equilibrium constraint as in \cite{berthelot2017began}, and (b) we use the autoencoder structure of the discriminator and pre-train it with the \textsl{real} UV targets prior to the adversarial training. Further details about the training procedure are presented in Section \ref{sec:training}. 

\textbf{Reconstruction loss.}
With the utilization of the adversarial loss \eqref{eq:adv_loss}, the generator $G$ is trying to ``fool'' the discriminator $D$. Nevertheless, this does \textsl{not} guarantee that the \textsl{fake} facial UV will be close to the corresponding \textsl{real}, target one. To impose this, we use an $L1$ loss between the \textsl{fake} sample $G\left(x\right)$ and the corresponding \textsl{real} one, $y$, so that they are as similar as possible, as in \cite{isola2017image}. Namely, the reconstruction loss is the following.
\begin{align}
\mathcal{L}_{rec} &= \mathbb{E}_{x}\norm{G\left(x\right) - y}_1.\label{eq:rec_loss} 
\end{align}

\textbf{Full objective.}
In sum, taking into account \eqref{eq:adv_loss} and \eqref{eq:rec_loss}, the full objective becomes
\begin{equation}\label{eq:full_loss}
\begin{array}{ll}
    \mathcal{L}_{D} = \mathbb{E}_{y}\left[\mathcal{L}\left(y\right)\right] - \lambda_{adv}\cdot\mathbb{E}_{x}\left[\mathcal{L}\left(G(x)\right)\right], \\
    \mathcal{L}_{G} = \mathbb{E}_{x}\left[\mathcal{L}\left(G(x)\right)\right] + \lambda_{rec}\cdot \mathcal{L}_{rec},
\end{array}
\end{equation}
\end{changemargin}
where $\lambda_{rec}$ is the hyper-parameter that controls how much emphasis should be put on the reconstruction loss. Overall, the discriminator $D$ tries to minimize $\mathcal{L}_D$ while the generator $G$ tries to minimize $\mathcal{L}_{G}$.

\subsection{Training procedure}\label{sec:training}
In this Section, we first describe how we pre-train the discriminator (autoencoder) $D$ and then provide details with respect to the adversarial training of 3DFaceGAN.

\begin{changemargin}{0cm}{0cm} 
\textbf{Pre-training the discriminator.} The majority of GANs in the literature utilize discriminator architectures with logit outputs that correspond to a prediction on whether the input fed into the discriminator is {\it real} or {\it fake}. Recently proposed GAN variations have nevertheless taken a different approach, namely by utilizing autoencoder structures as discriminators \Citep{zhao2016energy, berthelot2017began}. Using an autoencoder structure in the discriminator $D$ is of paramount importance in the proposed 3DFaceGAN. The benefit is twofold: (a) we can pre-train the autoencoder $D$ acting as discriminator prior to the adversarial training, which leads to better quantitative as well as more compelling visual results \footnote{note that pre-training $D$ is not possible when the outputs are logits since there are no fake data to compare against prior to the adversarial training.}, and (b) we are able to compute the actual UV space dense loss, as compared to simply deciding on whether the input is real or fake. As we empirically show in our experiments and ablation studies, this approach encourages the generator to produce more realistic results than other, state-of-the-art methodologies. 

\textbf{Adversarial training.}
Before starting the adversarial training, we initialize the weights and biases\footnote{for brevity in the text, we will use the term parameters to refer to the weights and the biases from this point onwards.} for both the generator $G$ and the discriminator $D$ utilizing the learned parameters estimated after the pre-training of $D$ (the architecture of $G$ is identical to the architecture of $D$). During the training phase of 3DFaceGAN, we freeze the parameter updates in the decoder parts for both the generator $G$ and the discriminator $D$. Furthermore, we utilize a low learning rate on the encoder and bottleneck parts of $G$ and $D$ so that overall the parameter updates are relatively close to the ones found during the pre-training of $D$.

\textbf{Network architectures}. The network architectures for both the discriminator $D$ and the generator $G$ are the same. In particular, each network is consisted of 2D convolutional blocks with kernel size of three, stride and padding size of one. Down-sampling is achieved by average 2D pooling with kernel and stride size of two. The convolution filters grow linearly in each down-sampling step. Up-sampling is implemented by nearest-neighbor with scale factor of two. The activation function that is primarily used is ELU \Citep{clevert2015fast}, apart from the last layer of both $D$ and $G$ where Tanh is utilized instead. At the bottleneck we utilize fully connected layers and thus project the tensors to a latent vector $b\in\mathbb{R}^{N_b}$. To generate more compelling visual results, we utilized skip connections \Citep{he2016deep, huang2017densely} in the first layers of the decoder part of both the generator and the discriminator. Further details about the network architectures are provided in Table \ref{table:architecture}.
 \end{changemargin}
 
 \renewcommand{\arraystretch}{1.5}
\begin{table*}[t]
\centering
\caption{Generator/Discriminator network architectures of 3DFaceGAN. As far as the notation is concerned, C denotes the number of input/output channels, K denotes the kernel size, S denotes the stride size, P denotes the padding size, AvgPool2D denotes average 2D pooling, UpNN denotes nearest-neighbor upsampling, and SF refers to the scaling factor size of the nearest-neighbor upsampling. CONV-BLOCK(C1, C2, K, S, P) and DECONV-BLOCK(C1, C2, K, S, P) refer to a block of two convolutions where the first is CONV(C1, C2, K, S, P) followed by an ELU \Citep{clevert2015fast} activation function and the second is CONV(C2, C2, K, S, P), also followed by an ELU \Citep{clevert2015fast} activation function.}
\vspace{0.3cm}
{\footnotesize
\label{table:architecture}
\begin{tabular}{c c c}
\hline
Part & Input $\rightarrow$ Output shape & Layer information\\
\hline
\hline
\multirow{7}{*}{Encoder} & (h, w, 3) $\rightarrow$ (h, w, n) & CONV-(Cn, K3x3, S1, P1), ELU \\
& (h, w, n) $\rightarrow$ ($\frac{h}{2}$, $\frac{w}{2}$, 2n) & CONV-BLOCK-(Cn, 2n, K3x3, S1, P1),  AvgPool2D(K2x2, S2) \\
 & ($\frac{h}{2}$, $\frac{w}{2}$, 2n) $\rightarrow$ ($\frac{h}{4}$, $\frac{w}{4}$, 3n) & CONV-BLOCK-(C2n, C3n, K3x3, S1, P1), AvgPool2D(K2x2, S2)\\
 & ($\frac{h}{4}$, $\frac{w}{4}$, 3n) $\rightarrow$ ($\frac{h}{8}$, $\frac{w}{8}$, 4n) & CONV-BLOCK-(C3n, C4n, K3x3, S1, P1), AvgPool2D(K2x2, S2)\\
 & ($\frac{h}{8}$, $\frac{w}{8}$, 4n) $\rightarrow$ ($\frac{h}{16}$, $\frac{w}{16}$, 5n) & CONV-BLOCK-(C4n, C5n, K3x3, S1, P1), AvgPool2D(K2x2, S2)\\
  & ($\frac{h}{16}$, $\frac{w}{16}$, 5n) $\rightarrow$ ($\frac{h}{32}$, $\frac{w}{32}$, 6n) & CONV-BLOCK-(C5n, C6n, K3x3, S1, P1), AvgPool2D(K2x2, S2)\\
  & ($\frac{h}{32}$, $\frac{w}{32}$, 6n) $\rightarrow$ ($\frac{h}{32}$, $\frac{w}{32}$, 6n) & CONV-BLOCK-(C6n, C6n, K3x3, S1, P1)\\
 \hline
 \hline
 Bottleneck$_1$ &  ($\frac{h}{32}\times\frac{w}{32}\times$6n) $\rightarrow$ n & Fully connected\\ 
 Bottleneck$_2$ & n $\rightarrow$ ($\frac{h}{32}\times\frac{w}{32}\times$n) & Fully connected\\ 
\hline
\hline
 \multirow{7}{*}{Decoder} & ($\frac{h}{32}$, $\frac{w}{32}$, n) $\rightarrow$ ($\frac{h}{16}$, $\frac{w}{16}$, n) & DECONV-BLOCK(Cn, Cn K3x3, S1, P1), UpNN(SF2) \\
 & ($\frac{h}{16}$, $\frac{w}{16}$, n) $\rightarrow$ ($\frac{h}{8}$, $\frac{w}{8}$, n) & DECONV-BLOCK(Cn, Cn, K3x3, S1, P1), UpNN(SF2) \\
 & ($\frac{h}{8}$, $\frac{w}{8}$, n) $\rightarrow$ ($\frac{h}{4}$, $\frac{w}{4}$, n) & DECONV-BLOCK(Cn, Cn, K3x3, S1, P1), UpNN(SF2) \\
 & ($\frac{h}{8}$, $\frac{w}{8}$, n) $\rightarrow$ ($\frac{h}{4}$, $\frac{w}{4}$, n) & DECONV-BLOCK(Cn, Cn, K3x3, S1, P1), UpNN(SF2) \\
  & ($\frac{h}{2}$, $\frac{w}{2}$, n) $\rightarrow$ (h, w, n) & DECONV-BLOCK(Cn, Cn, K3x3, S1, P1), UpNN(SF2) \\
   & (h, w, n) $\rightarrow$ (h, w, n) & DECONV-BLOCK(Cn, Cn, K3x3, S1, P1)\\
    & (h, w, n) $\rightarrow$ (h, w, 3) & DECONV(Cn, C3, K3x3, S1, P1), Tanh\\
 \hline
 \hline
\end{tabular}}
\end{table*}
 
\subsection{3D face generation}\label{sec:generation}
Variational autoencoders (VAEs) \Citep{kingma2013auto} are widely used for generating new data using autoencoder-like structures. In this setting, VAEs add a constraint on the latent embeddings of the autoencoders that forces them to roughly follow a normal distribution. We can then generate new data by sampling a latent embedding from the normal distribution and pass it to the decoder. Nevertheless, it was empirically shown that enforcing the embeddings in the training process to follow a normal distribution leads to generators that are unable to capture high frequency details \Citep{litany2017deformable}. To alleviate this, we propose to generate data using Algorithm \ref{algo:generate_data}, which better retains the generated data fidelity, as shown in Section \ref{sec:experiments}.

\subsection{3DFaceGAN for multi-label 3D data}\label{sec:multi_label}
Over the last few years, databases annotated with regards to multiple labels are becoming available in the scientific community. For instance, 4DFAB \Citep{cheng20184dfab} is a publicly available 3D facial database containing data annotated with respect to multiple expressions. 

We can extend 3DFaceGAN to handle data annotated with regards to multiple labels as follows. Without any loss of generality, suppose there are three labels in the database (e.g., expressions \textsl{neutral}, \textsl{happiness} and \textsl{surprise}). We adopt the so-called one-hot representation and thus denote the existence of a particular label in a datum by $1$ and the absence by $0$. For example, a 3D face datum annotated with the label \textsl{happiness} will have the following label representation: $l=[0, 1, 0]$, where the first entry corresponds to the label \textsl{neutral}, the second to the label \textsl{happiness} and the third to the label \textsl{surprise}. We then choose the desired $l$ we want to generate (e.g., if we want to translate a neutral face to a surprised one, we would choose $l=[0, 0, 1]$) and then spatially replicate it and concatenate it in the input that is then fed to the generator. The real target is the actual expression (in this case \textsl{surprise}) with the corresponding $l$ spatially replicated and concatenated. Apart from this change, the rest of the training process is exactly the same as the one described in Section \ref{sec:training}. 

Finally, to generate 3D facial data with respect to a particular label, we follow the same process as the one presented in Algorithm \ref{algo:generate_data}, with the only difference being that we extract different pairs of ($\boldsymbol{\mu}_Z$, $\boldsymbol{\Sigma}_Z$) for every subset of the data, each corresponding to a particular label in the database. We then choose the pair ($\boldsymbol{\mu}_Z$, $\boldsymbol{\Sigma}_Z$) corresponding to the desired label and sample from this multi-variate Gaussian distribution.
 
\begin{algorithm*}[t]
\DontPrintSemicolon 
\textbf{Step 1:} Train 3DFaceGAN utilizing \eqref{eq:full_loss}.\;
\textbf{Step 2:} Extract the trained $G$, and for all $N$ training facial UV maps:\;
\For{$i = 1:N$} {
Input UV map $\mathbf{x}_i$ in $G$.\; 
Extract the corresponding bottleneck $\mathbf{z}_i\in\mathbb{R}^{N_b\times 1}$.
}
\textbf{Step 3:} Concatenate column-wise all of the bottlenecks, i.e., $\mathbf{Z} = \left[\mathbf{z}_1, \mathbf{z}_2, \ldots, \mathbf{z}_N\right]$.\;
\textbf{Step 4:} Extract the mean $\boldsymbol{\mu}_Z$ of $\mathbf{Z}$ and the covariance $\boldsymbol{\Sigma}_Z$ of the zero-mean $\mathbf{Z}$.\;
\textbf{Step 5:} To generate new data, retain only the trained Bottleneck$_2$ and the Decoder part of $G$ (see Table \ref{table:architecture} for the network structures) and sample a new $\mathbf{z}_i$ (i.e., Bottleneck$_2$ input) from the multivariate Gaussian $\mathcal{N}\left(\boldsymbol{\mu}_Z,\boldsymbol{\Sigma}_Z\right)$.\;
\caption{3D face generation algorithm.}
\label{algo:generate_data}
\end{algorithm*}
 
\section{Experiments}
\label{sec:experiments}
In this Section we (a) describe the databases which we used to carry out the experiments utilizing 3DFaceGAN, (b) provide information with respect to the data preprocessing we conducted prior to feeding the 3D data into the network, (c) succinctly describe the baseline state-of-the-art algorithms we employed for comparisons and (d) provide quantitative as well as qualitative results on a series of experiments that demonstrate the superiority of 3DFaceGAN.

\subsection{Databases}

\subsubsection{The Hi-Lo database}
{\it Hi-Lo} database contains approximately $6,000$ 3D facial scans captured during a special exhibition in the Science Museum, London. It is divided into the high quality data (\emph{Hi}) recorded with a 3dMD face capturing system and the low quality (\emph{Lo}) data captured with a V1 Kinect sensor. All the subjects were recorded in neutral expression. The overlapping subjects that were recorded in both frameworks were approximately $3,000$.

The 3dMD apparatus utilizes a 4 camera structured light stereo system which can create 3D triangular surface meshes composed of approximately $60,000$ vertices joined into approximately $120,000$ triangles. Moreover, the low quality database was captured with a KinectFusion framework \Citep{newcombe2011kinectfusion}. In contrast to the 3dMD system, multiple frames are required to build a single 3D representation of the subject's face. The fused meshes were built by employing a $6,083$ voxel grid. In order to accurately reconstruct the entire surface of the faces, a circular motion scanning pattern was carried out. Each subject was instructed to stay still in a fixed pose during the entire scanning process with a neutral facial expression. The frame rate for every subject was constant at $8$ frames per second.

Furthermore, all $3,000$ subjects provided metadata about themselves, including their gender, age, and ethnicity. The database covers a wide variety of age, gender ($48\%$ male, $52\%$ female), and ethnicity ($82\%$ White, $9\%$ Asian, $5\%$ Mixed Heritage, $3\%$ Black and $1\%$ other).

{\it Hi-Lo} database was utilized for the experiments of 3D face representation and generation, where we utilized the high quality data to train 3DFaceGAN. Moreover, {\it Hi-Lo} database was used for demonstrating the capabilities of 3DFaceGAN in a 3D face translation setting, where the low quality data are translated into high quality ones. In all of the training tasks, $85\%$ of the data were used for training and the rest were used for testing.

\subsubsection{4DFAB database}
4DFAB database \Citep{cheng20184dfab} contains 3D facial data from $180$ subjects ($60$ females, $120$ males), aged from $5$ to $75$ years old. The subjects vary in their ethnicity background, coming from more than $30$ different ethnic groups. For the capturing process, the DI4D dynamic capturing system \footnote{http://www.di4d.com} was used. 

4DFAB \Citep{cheng20184dfab} contains data varying in expressions, such as {\it neutral}, {\it happiness}, and {\it surprise}. As a result, we utilized it to showcase 3DFaceGAN's capability in successfully handling data annotated with multiple labels in the task of 3D face translation as well as generation. In all of the training tasks, $85\%$ of the data were used for training and the rest were used for testing.

\subsection{Data preprocessing}\label{sec:data_preprocessing}
In order to feed the 3D data into a deep network several steps need to be carried out. Since we employ various databases, the representation of the facial topology is not consistent in terms of vertex number and triangulation. To this end, we need to find a suitable template $T$ that can easily retain the information of all raw scans across all databases and describe them with the same triangulation/topology. We utilized the mean face mesh of the LSFM model proposed by \cite{booth20163d}, which consists of approximately $54,000$ vertices that are sufficient to capture high frequency facial details. We then bring the raw scans in dense correspondence by morphing non-rigidly the template mesh to each one of them. For this task, we utilize an optimal-step Non-rigid Iterative Closest Point algorithm \Citep{de2010optimal} in combination with a per vertex weighting scheme. We weight the vertices according to the Euclidean distance measured from the tip of the nose. The greater the distance from the nose tip, the bigger the weight that is assigned to that vertex, i.e., less flexible to deform. In that way we are able to avoid the noisy information recorded by the scanners on the outer regions of the raw scans.

Following the analysis of the various methods of feeding 3D meshes in deep networks in Section \ref{sec:3D_rep}, we chose to describe the 3D shapes in the UV domain. UV maps are usually utilized to store texture information. In our case, we store the spatial location of each vertex as an RGB value in the UV space. In order to acquire the UV pixel coordinates for each vertex, we start by unwrapping our mesh template $T$ into a 2D flat space by utilizing an optimal cylindrical unwrapping technique proposed by \cite{booth2014optimal}. Before storing the 3D coordinates into the UV space, all meshes are aligned in the 3D spaces by performing the General Procrustes Analysis \Citep{gower1975generalized} and are normalized to be in the scale of $[1,-1]$. Afterwards, we place each 3D vertex in the image plane given the respective UV pixel coordinate. Finally, after storing the original vertex coordinates, we perform a 2D nearest point interpolation in the UV domain to fill out the missing areas in order to produce a dense representation of the originally sparse UV map. Since the number of vertices in $S_T$ is more than $50K$, we choose a $256\times 256\times 3$ tensor as the UV map size, which assists in retrieving a high precision point cloud with negligible re-sampling errors. A graphical representation of the preprocessing pipeline can be seen in Figure~\ref{fig:_preprocessing_uvs}.

\begin{figure*}[t]
\begin{center}
\includegraphics[width=0.9\linewidth]{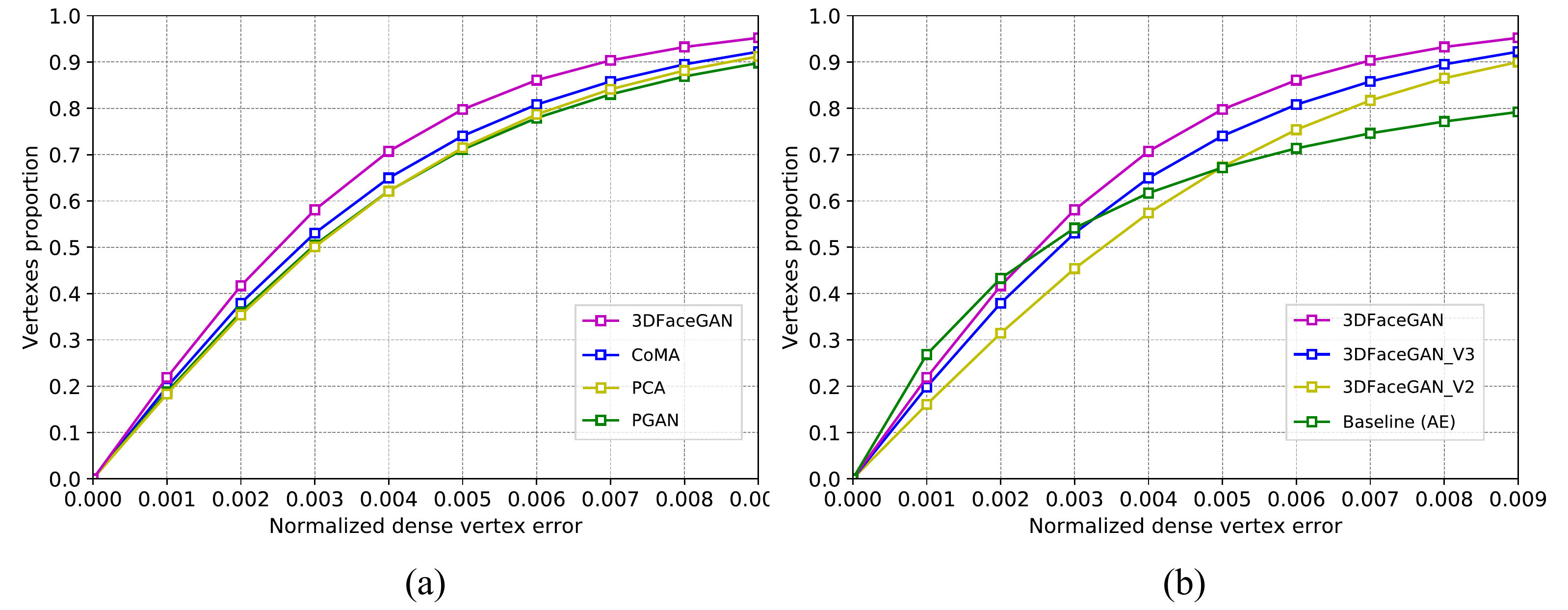}
\caption{(a) Generalization results on the test set for the 3D face representation task. The results are presented as cumulative error distributions of the normalized dense vertex errors. 3DFaceGAN outperforms all of the compared methods by a large margin. (b) Ablation study generalization results for the 3D face representation task. The results are presented as cumulative error distributions of the normalized dense vertex errors.}
\label{fig:error_plot}
\end{center}
\end{figure*}

\begin{table}[t]
\caption{Generalization metric for the meshes of the test set for the 3D face representation task. The table reports the mean error (Mean), the standard deviation (std), the Area Under the Curve (AUC), and the Failure Rate (FR) of the Cumulative Error Distributions of Fig. \ref{fig:error_plot}a.}
\label{tab:error_gen}
\vspace{0.3cm}
\centering
\begin{tabular}{|l|cccc|}
\hline
\emph{Method} & \emph{Mean} & \emph{std} & \emph{AUC} & \emph{FR (\%)} \\
\hline\hline
\textbf{3DFaceGAN} & 0.0031 & \textpm 0.0028 & \textbf{0.741} & \textbf{1.42e-7} \\
CoMA & 0.0038 & \textpm 0.0037 & 0.716 & 3.66e-7 \\
PCA & 0.0040 & \textpm 0.0040 & 0.711 & 0.91e-6 \\
PGAN & 0.0041 & \textpm 0.0041 & 0.705 & 1.22e-6 \\
\hline
\end{tabular}
\end{table}

\begin{table}[t]
\caption{Ablation study generalization results for the 3D face representation task. The table reports the Area Under the Curve (AUC) and Failure Rate (FR) of the Cumulative Error Distributions of Fig. \ref{fig:error_plot}b.}
\label{tab:error_ablation}
\vspace{0.3cm}
\centering
\begin{tabular}{|l|cc|}
\hline
\emph{Method} & \emph{AUC} & \emph{FR (\%)} \\
\hline\hline
\textbf{3DFaceGAN} & \textbf{0.741} & \textbf{1.42e-7} \\
3DFaceGAN\_V3  & 0.736 & 2.62e-7 \\ 
3DFaceGAN\_V2 & 0.704 & 3.15e-6 \\
Baseline (AE) & 0.697 & 4.24-6 \\
\hline
\end{tabular}
\end{table}

\begin{figure*}[t]
\begin{center}
\includegraphics[width=0.9\linewidth]{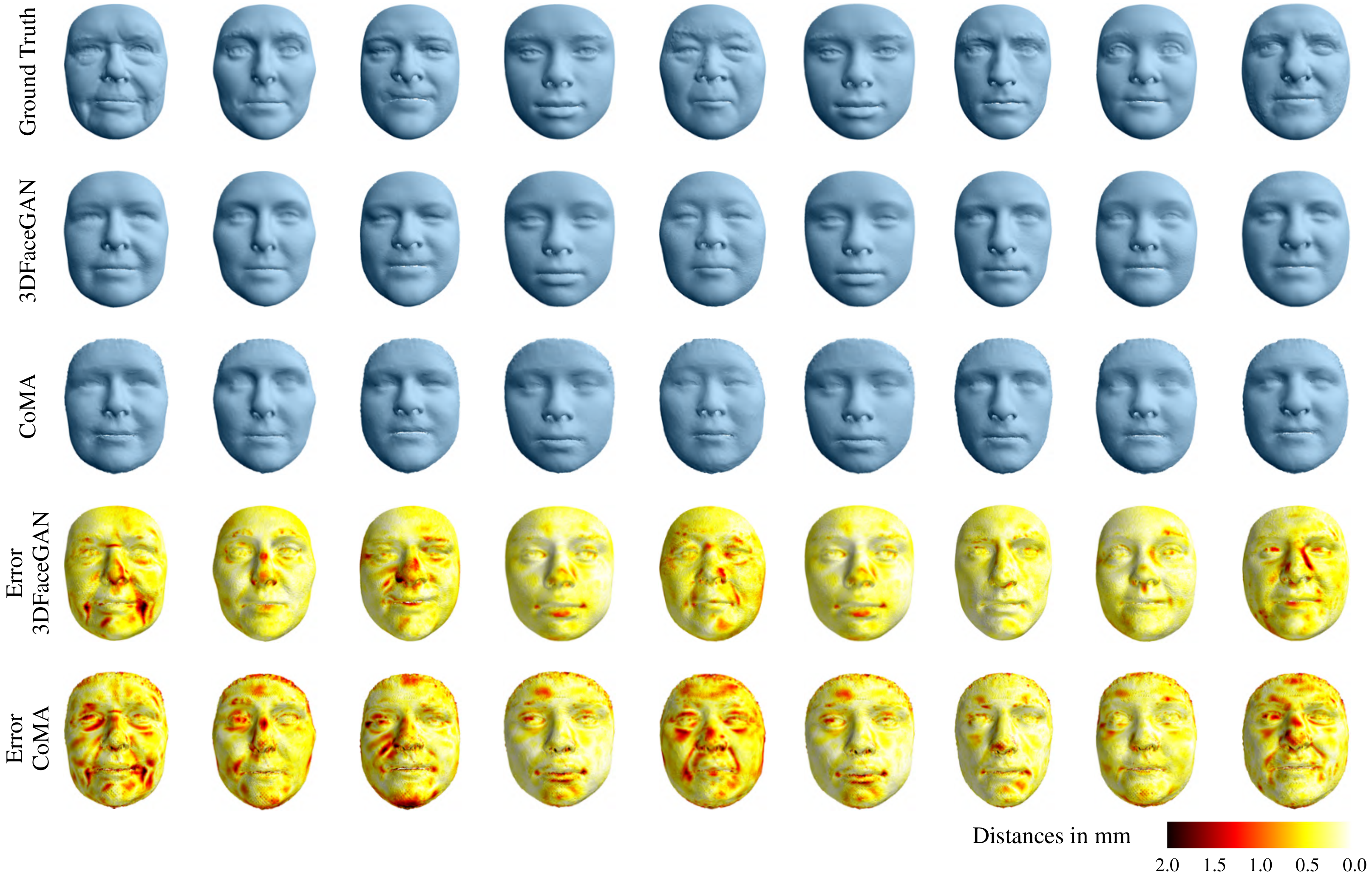}
\caption{Qualitative results of 3DFaceGAN compared to CoMA \Citep{ranjan2018generating} in the 3D representation task. Moreover, heatmaps are provided, visualizing the errors of both approaches against the ground truth test data. As evidenced, 3DFaceGAN is able to better capture the variation in the test data, especially in the eye and nose regions, where most of the non-linearities are present.}
\label{fig:qualitative}
\end{center}
\end{figure*}

\begin{figure}[t]
\begin{center}
\includegraphics[width=0.8\linewidth]{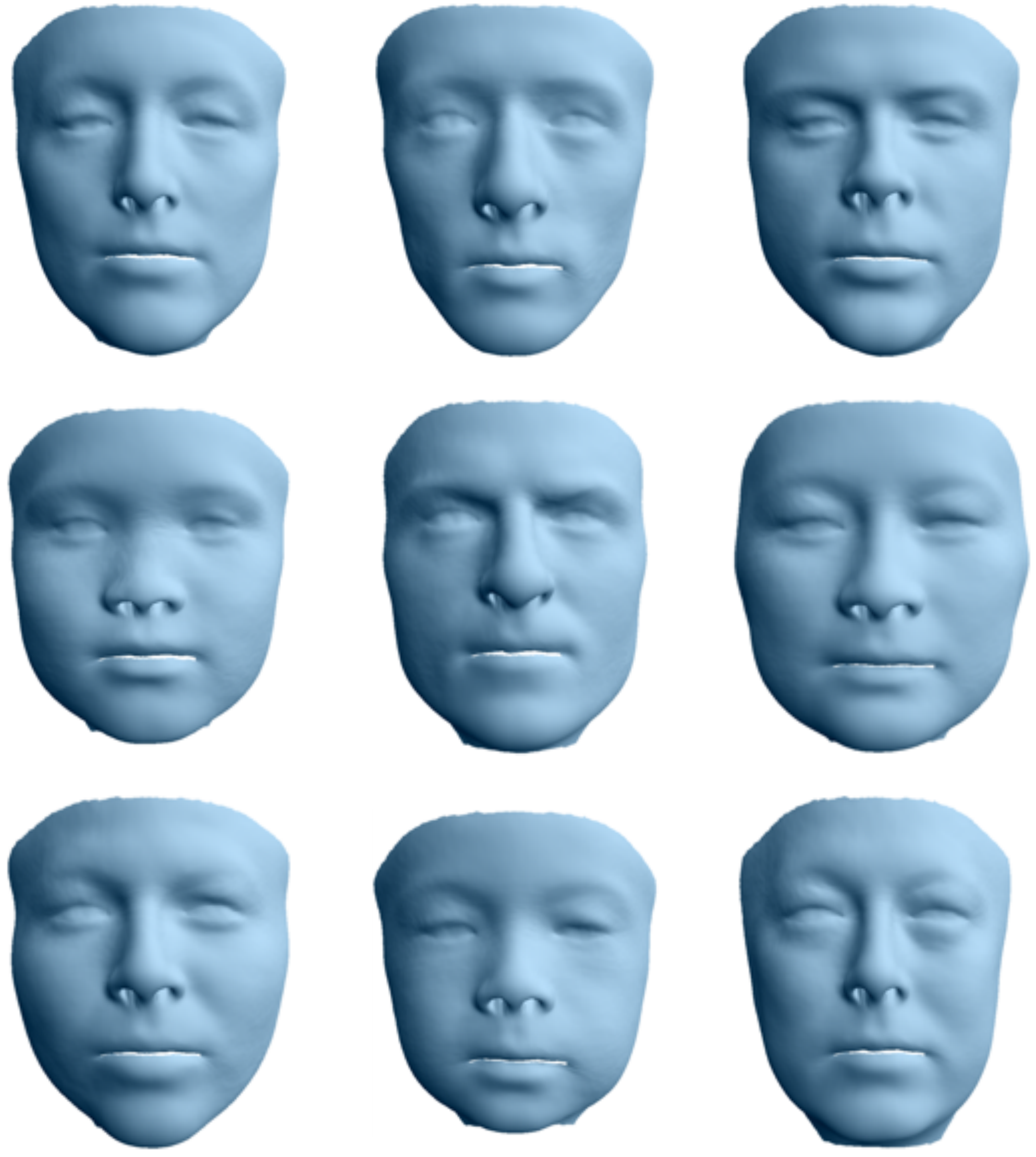}
\caption{Generated faces utilizing 3DFaceGAN.}
\label{fig:generations}
\end{center}
\end{figure}

\begin{figure*}[ht]
\begin{center}
\includegraphics[width=0.82\linewidth]{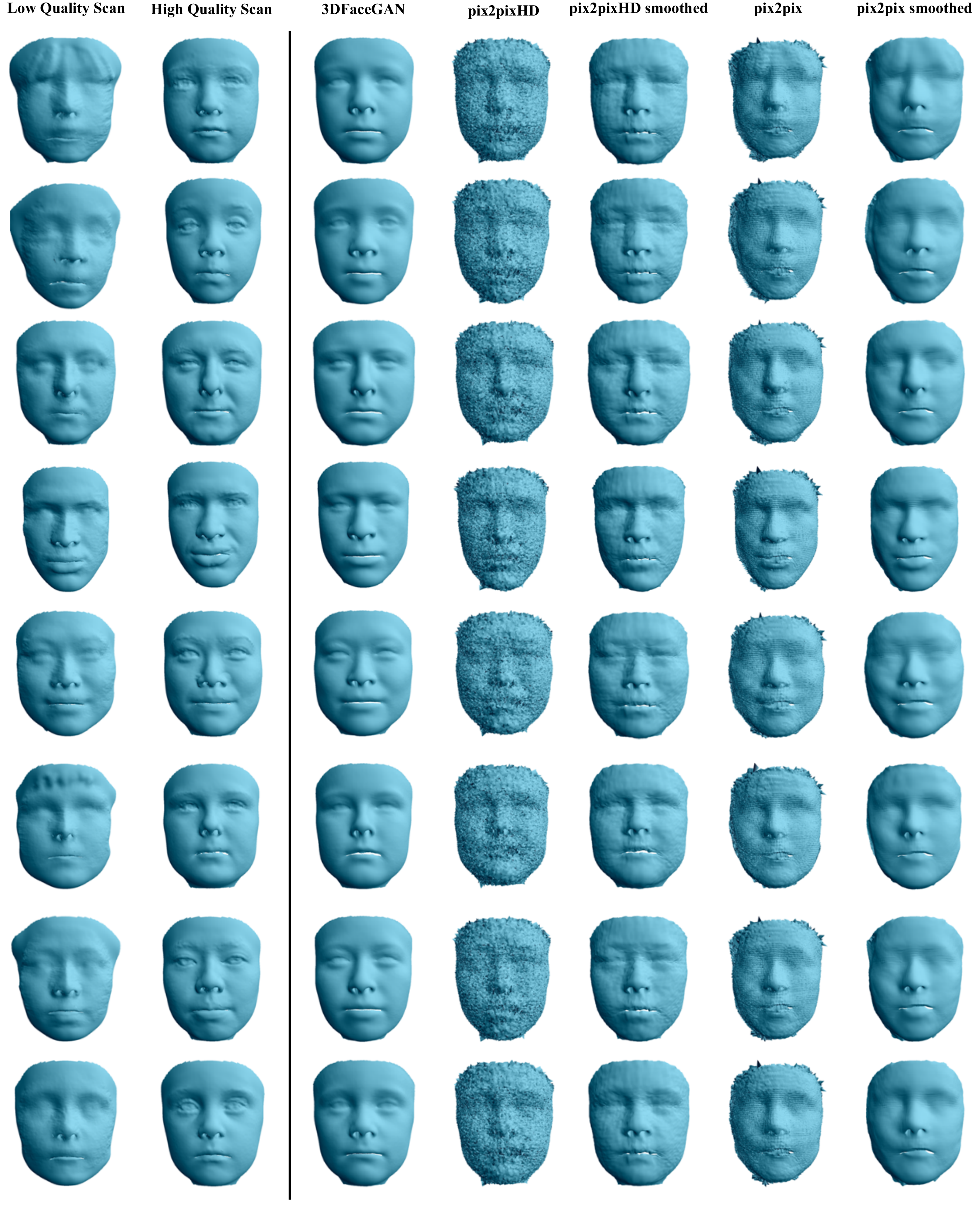}
\caption{The qualitative results of our approach compared to state-of-the-art baseline GAN methods in the 3D face translation task. The first column depicts the low quality input mesh whereas the second column represent the high quality ground truth meshes. We depict the raw results of pix2pixHD \Citep{wang2018high} and pix2pix \Citep{isola2017image} along with their smoothed versions. As a smoothing technique we utilized a standard Laplacian smoothing operator.}
\label{fig:qualitative}
\end{center}
\end{figure*}

\subsection{Training}
We trained all 3DFaceGAN models utilizing Adam \Citep{kingma2014adam} with $\beta_1=0.5$ and $\beta_2=0.999$. The batch size we used for the pre-training of the discrminator was $32$ for a total of $300$ epochs. The batch size we used for 3DFaceGAN was $16$ for a total of $300$ epochs. For our model we used $n=128$ convolution filters and a bottleneck of size $b=128$. The total number of trainable parameters was $38.5\times 10^6$. The learning rates that we used for both the pre-training and training of the discriminator was $5e-5$ and the same was for the training of the generator. We linearly decayed the learning rate by $5\%$ every $30$ epochs during training. For the rest of the parameters, we used $\lambda_{adv}=1e-3$, $\lambda_{rec}=1$. Overall training time on a GV100 NVIDIA GPU was about $5$ days.

\subsection{3D Face Representation}\label{sec:representation}
In the 3D face representation (reconstruction) experiments, we utilize the high quality 3D face data from the {\it Hi-Lo} database to train the algorithms. In particular, we feed the high quality 3D data as inputs to the models and use the same data as target outputs. Before providing the qualitative as well as quantitative results, we briefly describe the baseline models we compared against as well as provide information about the error metric we used for the quantitative assessment.

\subsubsection{Baseline models}
In this Section we briefly describe the state-of-the-art models we utilized to compare 3DFaceGAN against. 

\subsubsection*{Vanilla Autoencoder (AE)}
Vanilla Autoencoder follows exactly the same structure of the discriminator we used in 3DFaceGAN. We used the same values for the hyper-parameters and the same optimization process. This is the main baseline we compared against and the results are provided in the ablation study in Section \ref{sec:ablation_rep}.

\subsubsection*{Convolutional Mesh Autoencoder (CoMA)}
In order to train CoMA \Citep{ranjan2018generating}, we use the authors' publicly available implementation and utilize the default parameter values, the only difference being that the bottleneck size is $128$, to make a fair comparison against 3DFaceGAN, where we also used a bottleneck size of $128$. 

\subsubsection*{Principal Component Analysis (PCA)}
We employ and train a standard PCA model \Citep{jolliffe2011principal} based on the meshes of our database we used for training. We aimed at retaining the $98\%$ of variance of our available training data which corresponds to the first $50$ principal components.

\subsubsection*{Progressive GAN (PGAN)}
In order to train PGAN \Citep{karras2017progressive}, we used the authors' publicly available implementation with the default parameter values. After the training is complete, in order to represent a test 3D datum, we \textsl{invert} the generator $G$ as in \cite{lucic2017gans} and \cite{mahendran2015understanding}, i.e., we solve $z^{*} = \argmin\norm{x - G(z)}$ by applying gradient descent on $z$ while retaining $G$ fixed \Citep{mahendran2015understanding}. 

\begin{figure*}[t]
\begin{center}
\includegraphics[width=0.9\linewidth]{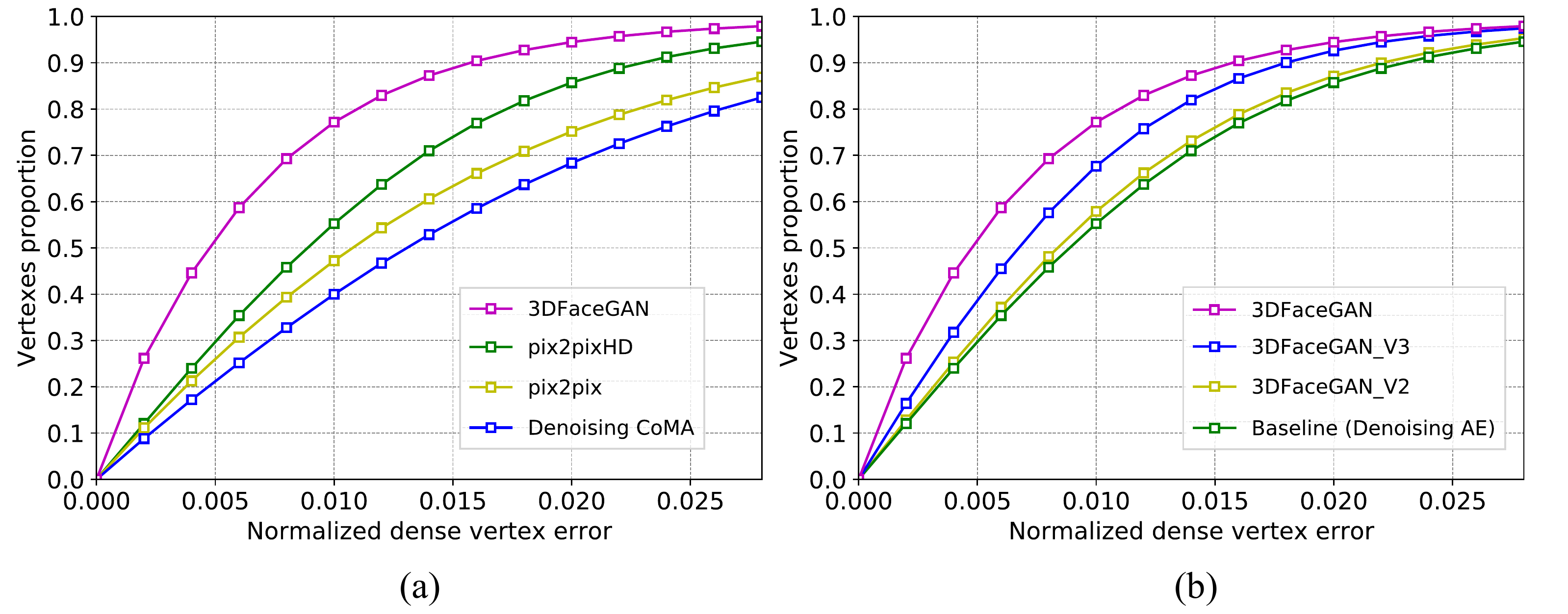}
\caption{(a) High quality estimation results for the 3D face translation task. The results are presented as cumulative error distributions of the normalized dense vertex errors. 3DFaceGAN outperforms all of the compared methods by a large margin. (b) Ablation study with respect to the 3D face translation task. The results are presented as cumulative error distributions of the normalized dense vertex errors.}
\label{fig:error_plot_translation}
\end{center}
\end{figure*}

\begin{table}[t]
\caption{High quality 3DRMSE results for the 3D face translation task. The table reports the Area Under the Curve (AUC) and Failure Rate of the Cumulative Error Distributions of Fig. \ref{fig:error_plot_translation}a.}
\vspace{0.3cm}
\centering
\begin{tabular}{|l|cc|}
\hline
\emph{Method} & \emph{AUC} & \emph{Failure Rate (\%)} \\
\hline\hline
\textbf{3DFaceGAN} & \textbf{0.827} & \textbf{5.49e-6} \\
pix2pixHD & 0.760 & 5.18e-5 \\
pix2pix & 0.757 & 1.81e-5 \\
Denoising CoMA & 0.742 & 2.41e-4 \\
\hline
\end{tabular}
\vspace{0.4cm}
\label{tab:dense_fit_error_1}
\end{table}

\begin{table}[t]
\caption{Ablation study 3DRMSE results for the 3D face translation task. The table reports the Area Under the Curve (AUC) and Failure Rate of the Cumulative Error Distributions of Fig. \ref{fig:error_plot_translation}b.}
\vspace{0.3cm}
\centering
\begin{tabular}{|l|cc|}
\hline
\emph{Method} & \emph{AUC} & \emph{Failure Rate (\%)} \\
\hline\hline
\textbf{3DFaceGAN} & \textbf{0.827} & \textbf{5.49e-6} \\
3DFaceGAN\_V3 & 0.819 & 8.70e-6 \\
3DFaceGAN\_V2 & 0.794 & 1.38e-5 \\
Baseline (Denoising AE) & 0.758 & 1.95e-5 \\
\hline
\end{tabular}
\label{tab:ablation_fit_error_1}
\end{table}

\subsubsection{Error metric}
A common practice when it comes to evaluating statistical shape models is to estimate the intrinsic characteristics, such as the {\it generalization} of the model \Citep{davies2008statistical}. The {\it generalization} metric captures the ability of a model to represent {\it unseen} 3D face shapes during the testing phase. Table \ref{tab:error_gen} presents the generalization metric for 3DFaceGAN compared against the baseline models. In order to compute the generalization error for a given model, we compute the per-vertex Euclidean distance between every sample of the test set and its corresponding reconstruction. We observe that the model which holds the best error results and thus demonstrates greater generalization capabilities is the proposed 3DFaceGAN with mean error $0.0031$ and standard deviation $0.0028$. Additionally, as shown in  Fig. \ref{fig:error_plot}a, which depicts the cumulative error distribution of the normalized dense vertex erors, 3DFaceGAN outperforms all of the baseline models.

\subsubsection{Ablation study}\label{sec:ablation_rep}
In this ablation study we investigate the importance of pre-training the discriminator $D$ prior to the adversarial training of 3DFaceGAN as well as the freezing of the weights in the decoder parts of both $D$ and $G$. More specifically, we compare 3DFaceGAN against the Vanilla Autoencoder (AE) and another two 3DFaceGAN possible variations, namely (a) the simplest case, where the discriminator and generator structures are retained as is, but \textsl{no} pre-training takes place prior to the adversarial training (we refer to this methodology as \textsl{3DFaceGAN\_V2}), (b) the case where (i) the discriminator and generator structures are retained as is, (ii) we pre-train the discriminator and initialize both the generator and the discriminator with the learned weights with \textsl{no} parameters frozen during the adversarial training (we refer to this methodology as \textsl{3DFaceGAN\_V3}). As shown in Fig. \ref{fig:error_plot}b and Table \ref{tab:ablation_fit_error_1}, 3DFaceGAN outperforms Vanilla AE and 3DFaceGAN\_V2 by a large margin. Moreover, 3DFaceGAN also outperforms 3DFaceGAN\_V3. As a result, not only does 3DFaceGAN have the best performance among the compared 3DFaceGAN variants, but it also requires less training time compared to 3DFaceGAN\_V3, as the parameters in the decoder parts of both the generator and the discriminator are not updated during the training phase and thus need not be computed. 

\subsection{3D Face Translation}\label{sec:face_translation}
In the 3D face translation experiments, we utilize the low and high quality 3D face data from the {\it Hi-Lo} database to train the algorithms. In particular, we feed the low quality 3D data as inputs to the models and use the high quality data as target outputs.

Before providing the qualitative as well as quantitative results, we briefly describe the baseline models we compared against as well as provide information about the error metric we used for the quantitative assessment. 

\subsubsection{Baseline models}\label{sec:translation}
In this Section we briefly describe the state-of-the-art deep models we utilized to compare 3DFaceGAN against. 

\subsubsection*{Denoising Vanilla Autoencoder (Denoising AE)}
Denoising Vanilla Autoencoder follows exactly the same structure as the Vanilla AE in Section \ref{sec:representation}, the only difference being the inputs fed to the network. This is the main baseline we compared against and the results are provided in the ablation study in Section \ref{sec:ablation_trans}.

\subsubsection*{Denoising Convolutional Mesh Autoencoder (Denoising CoMA)}
Denoising CoMA \Citep{ranjan2018generating}, follows exactly the same structure as the Vanilla AE in Section \ref{sec:representation}, the only difference being again the inputs fed to the network.
 
\subsubsection*{pix2pix}
pix2pix \Citep{isola2017image} is amongst the most widely utilized GANs for image to image translation applications. We used the official implementation and hyper-parameter initializations provided by the authors in \Citep{isola2017image}.

\begin{figure}[t]
\begin{center}
\includegraphics[width=1\linewidth]{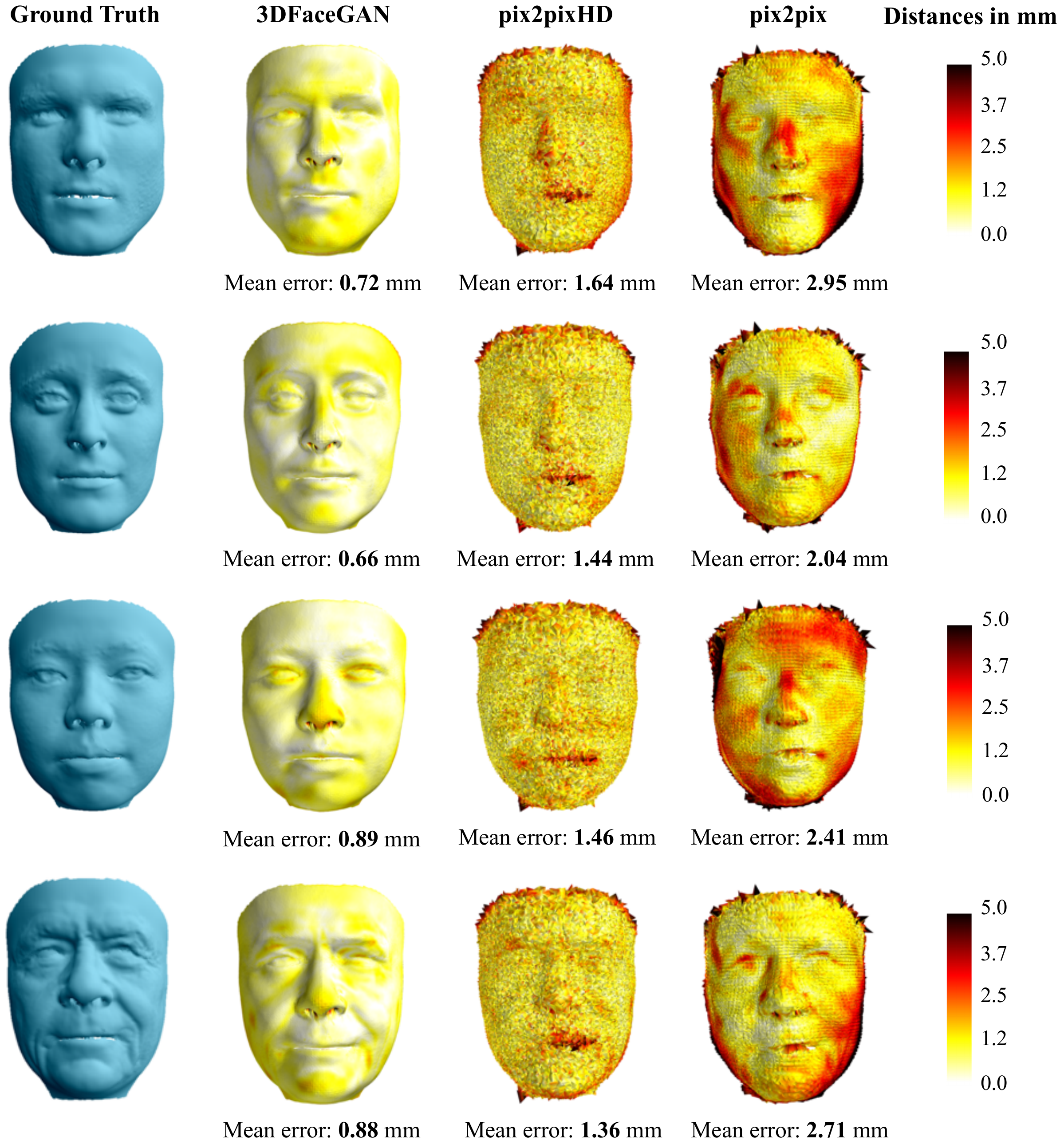}
\caption{Reconstruction quality of our proposed GAN network along with pix2pixHD \Citep{wang2018high} and pix2pix \Citep{isola2017image} in the 3D face translation task. As it can be seen, the mean error of 3DFaceGAN is considerably less than the other two approaches.}
\label{fig:heat_maps}
\end{center}
\end{figure}

\subsubsection*{pix2pixHD}
More recently pix2pixHD \Citep{wang2018high} was proposed, which can be considered as an extension of pix2pix \Citep{isola2017image} and which is able to better handle data of higher resolution. We used the official implementation and hyper-parameter initializations provided by the authors in \cite{wang2018high}. As evinced in Fig. \ref{fig:qualitative}, Fig. \ref{fig:error_plot_translation}, and Fig. \ref{fig:heat_maps}, pix2pixHD \Citep{wang2018high} outperforms pix2pix \Citep{isola2017image}, and this is expected since pix2pixHD \Citep{wang2018high} uses more intricate structures for both the generator and discriminator networks. 

\subsubsection{Error metric}
For each low quality test mesh we aim to estimate the high quality representation based on the 3dMD ground truth data. The error metric between the estimated and the real high quality mesh is a standard 3D Root Mean Square Error (3DRMSE) where the Euclidean distances are computed between the two meshes and normalized based on the inter-ocular distance of the test mesh. Before computing the metric error we perform dense alignment between each test mesh and its corresponding ground truth by implementing an iterative closest point (ICP) algorithm \Citep{besl1992method}. In order to avoid any inconsistencies in the alignment we compute a point-to-plain rather than a point-to-point error. Finally, the measurements are performed in the inner part of the face, where we crop each test mesh at a radius of $150$\emph{mm} around the tip of the nose. As can be clearly seen in Fig. \ref{fig:error_plot_translation}a as well as in Table \ref{tab:dense_fit_error_1}, 3DFaceGAN outperforms all of the compared state-of-the-art methods.

\subsubsection{Ablation study}\label{sec:ablation_trans}
For the ablation study in this set of experiments, we use exactly the same 3DFaceGAN variants as the ones we utilized in Section \ref{sec:ablation_rep}. Moreover, instead of the vanilla AE in this experiment we utilize the denoising AE. As evinced in Fig. \ref{fig:error_plot_translation}b and Table \ref{tab:ablation_fit_error_1}, 3DFaceGAN clearly outperforms all of the compared models.

\begin{figure*}[t]
\begin{center}
\includegraphics[width=1\linewidth]{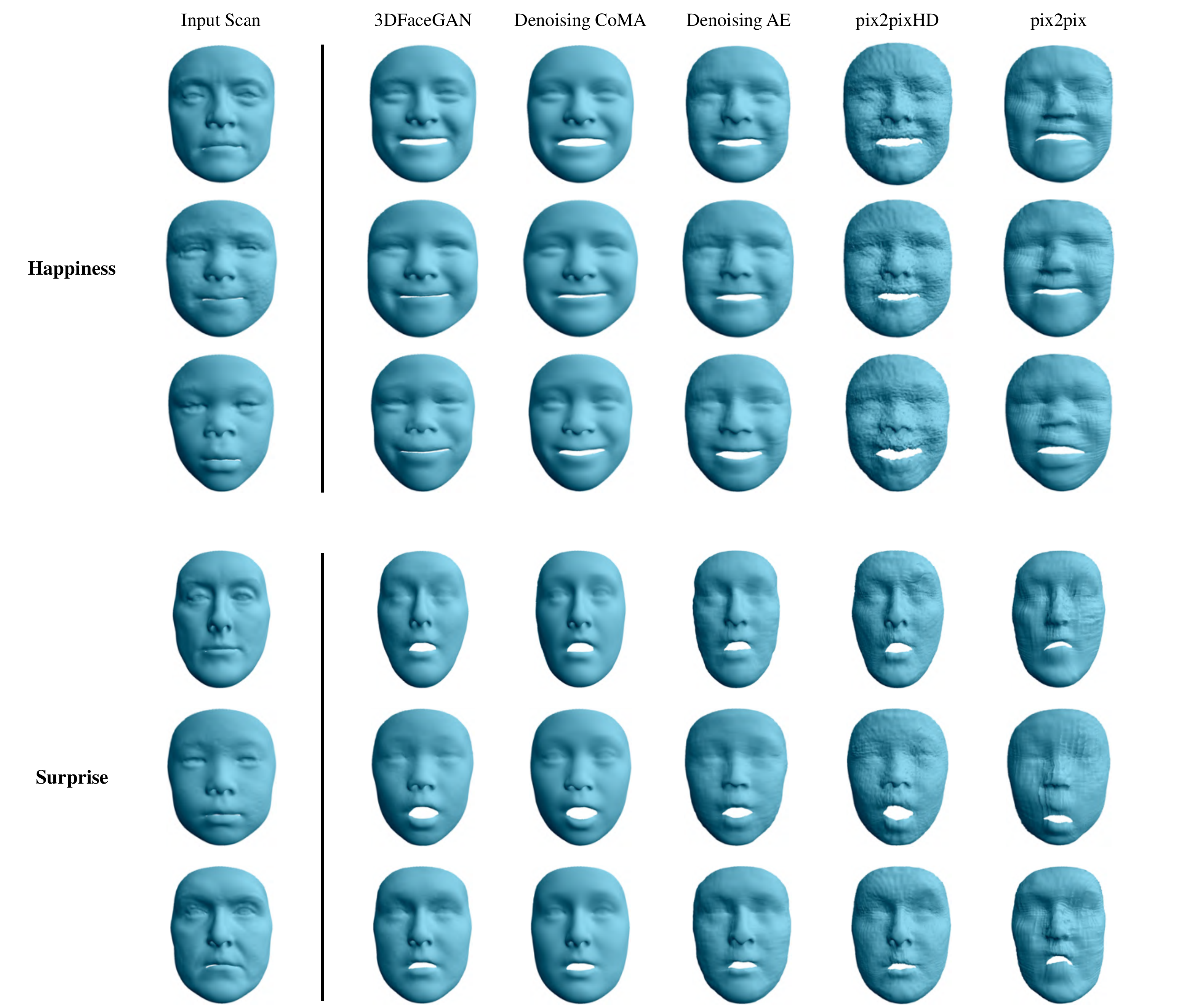}
\caption{Qualitative results of our approach compared to state-of-the-art baseline GAN methods in the multi-label 3D face translation task in various expressions (e.g., happiness, surprise) trained with the 4DFAB \Citep{cheng20184dfab} database. The first column depicts the neutral input mesh whereas the rest of the columns represent the translated meshes of the respective state-of-the art methods compared to our approach. As can be seen, 3DFaceGAN is able to retain the high-frequency details in a higher level compared to CoMA \Citep{ranjan2018generating}, the second best method, which produces more smoothed outputs.}
\label{fig:expression_qual}
\end{center}
\end{figure*}

\subsection{Multi-label 3D Face Translation}\label{sec:multi_label_translation}
 In this experiment we utilize 4DFAB \Citep{cheng20184dfab} for the multi-label transfer of expressions. In particular, we feed the \textsl{neutral} faces to the models and receive as outputs either the ones bearing the label \textsl{happiness} or \textsl{surprise}. It should be noted here that whereas 3DFaceGAN requires only a single model to be trained under the multi-label expression translation scenario, the rest of the compared models require different trained models for each label, i.e., a model for expression \textsl{happiness} and a model for expression \textsl{surprise}. As baseline models for comparisons, we use exactly the same as the ones in Section \ref{sec:face_translation}, the only difference being the inputs fed to network as well as the corresponding targets. Qualitative comparisons against the compared methods are presented in Fig. \ref{fig:expression_qual}.

\subsection{3D Face Generation}
In the 3D face generation experiment, we utilized the high quality data of the \textsl{Hi-Lo} database to train the algorithms. In particular, we feed the high quality 3D data as inputs to the models and use the same data as target outputs. 

\subsubsection{Baseline models}
The baseline models we used in this set of experiments are the same as the ones presented in Section \ref{sec:representation}.

\begin{table}[t]
\caption{Specificity metric on the test set for the 3D face generation task. We generate $10,000$ random faces from each model. The table reports the mean error (Mean) and the standard deviation (std).}
\label{tab:specificity}
\vspace{0.3cm}
\centering
\begin{tabular}{|l|cc|}
\hline
\emph{Method} & \emph{Mean} & \emph{std} \\
\hline\hline
\textbf{3DFaceGAN} & \textbf{1.28} & \textpm
 \textbf{0.183} \\
CoMA & 1.40 & \textpm 0.205 \\
PCA  & 1.43 & \textpm 0.232 \\
PGAN & 1.79 & \textpm 0.189 \\
\hline
\end{tabular}
\end{table}

\begin{figure}[t]
\begin{center}
\includegraphics[width=0.8\linewidth]{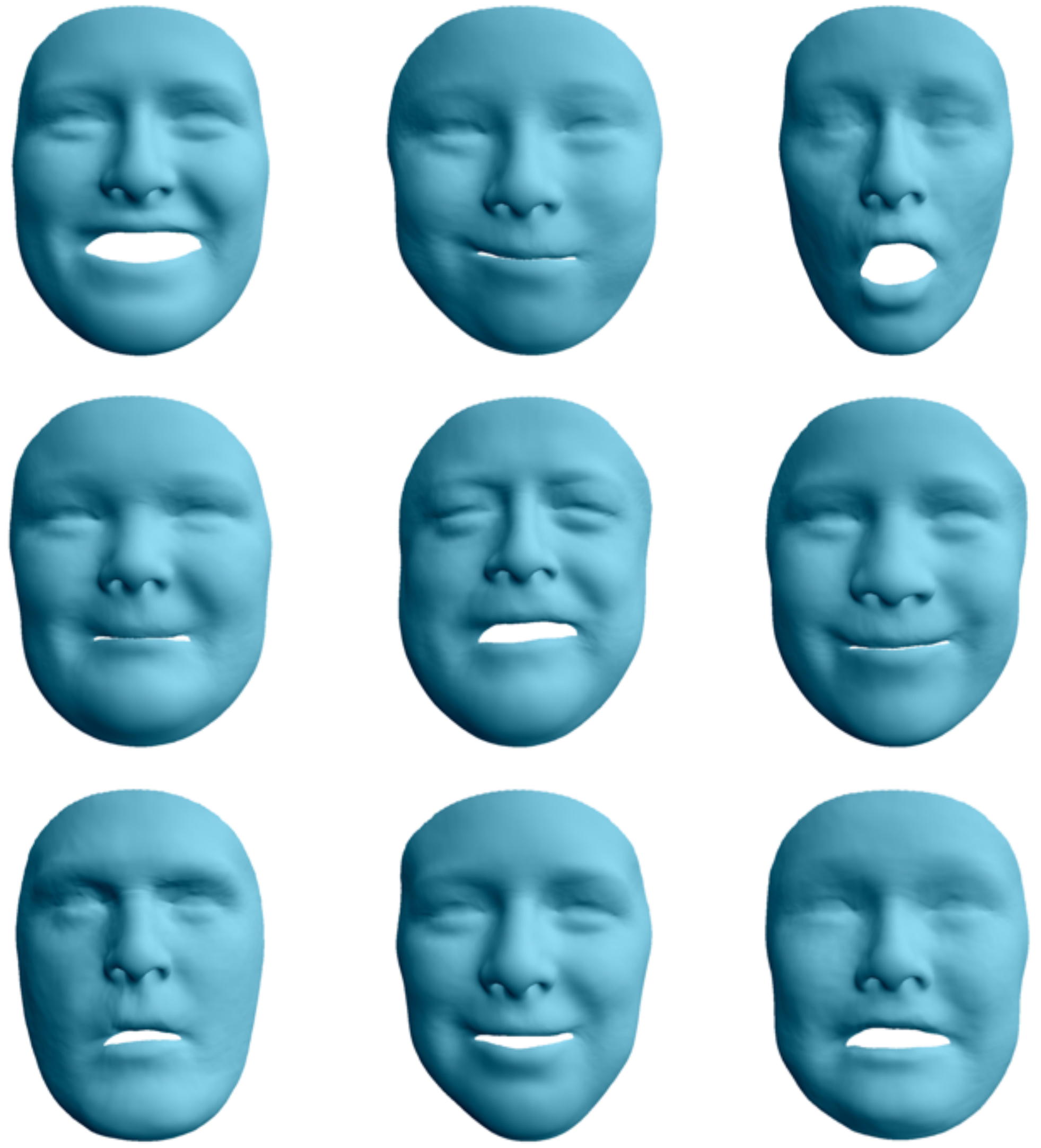}
\caption{Generated faces with expression utilizing 3DFaceGAN multi-label approach.}
\label{fig:gen_expressions}
\end{center}
\end{figure}

\subsubsection{Error metric}
The metric of choice to quantitatively assess the performance of the models in this set of experiments is {\it specificity} \Citep{brunton2014review}. For a randomly generated 3D face, {\it specificity} metric measures the distance of this 3D face to its nearest real 3D face belonging in the test, in terms of minimum per vertex distance over all samples of the test set. To evaluate this metric, we randomly generate $n=10,000$ face meshes from each model. Table \ref{tab:specificity} reports the specificity metric for 3DFaceGAN compared against the baseline models. In order to generate random meshes utilizing 3DFaceGAN, we sample from a multivariate Gaussian distribution, as explained in Section \ref{sec:generation}. To generate random meshes utilizing PGAN \Citep{karras2017progressive}, we sample new latent embeddings from the multivariate normal distribution and feed them to the generator $G$. To generate random faces utilizing CoMA \Citep{ranjan2018generating}, we utilize the proposed variational convolutional mesh autoencoder structure, as described in \Citep{ranjan2018generating}. For the PCA model \Citep{jolliffe2011principal}, we generate meshes directly from the latent eigenspace by drawing random samples from a Gaussian distribution defined by the principal eigenvalues. As shown in Table \ref{tab:specificity}, 3DFaceGAN achieves the best specificity error, outperforming all compared methods by a large margin. 

In Fig. \ref{fig:generations}, we present various visualizations of realistic 3D faces generated by 3DFaceGAN. As can be clearly seen, 3DFaceGAN is able to generate data varying in ethnicity, age, etc., thus capturing the whole population spectrum.

\subsection{Multi-label 3D Face Generation}
In this set of experiments, we utilized the 4DFAB \Citep{cheng20184dfab} data to generate random subjects of various expressions such as \textsl{happiness} and \textsl{surprise}, as seen in Fig. \ref{fig:gen_expressions}. The 3D faces were generated utilizing the methodology detailed in Section \ref{sec:multi_label}. As evinced, 3DFaceGAN is able to generate expressions of subjects varying in age and ethnicity, while retaining the high-frequency details of the 3D face. 

\section{Conclusion}
In this paper we presented the first GAN tailored for the tasks of 3D face representation, generation, and translation. Leveraging the strengths of autoencoder-based discriminators in an adversarial framework, we propose 3DFaceGAN, a novel technique for training on large-scale 3D facial scans. As shown in an extensive series of quantitative as well as qualitative experiments against other state-of-the-art deep networks, 3DFaceGAN improves upon state-of-the-art algorithms for the tasks at-hand by a significant margin.

\begin{acknowledgements}
Stylianos Moschoglou is supported by an EPSRC DTA studentship from Imperial College London, Stylianos Ploumpis by the EPSRC Project EP/N007743/1 (FACER2VM), and Stefanos Zafeiriou by the EPSRC Project EP/S010203/1 (DEFORM).
\end{acknowledgements}

{\small
\bibliographystyle{spbasic}
\bibliography{egbib}

\begin{thebibliography}{51}
\providecommand{\natexlab}[1]{#1}
\providecommand{\url}[1]{{#1}}
\providecommand{\urlprefix}{URL }
\expandafter\ifx\csname urlstyle\endcsname\relax
  \providecommand{\doi}[1]{DOI~\discretionary{}{}{}#1}\else
  \providecommand{\doi}{DOI~\discretionary{}{}{}\begingroup
  \urlstyle{rm}\Url}\fi
\providecommand{\eprint}[2][]{\url{#2}}

\bibitem[{Berthelot et~al.(2017)Berthelot, Schumm, and
  Metz}]{berthelot2017began}
Berthelot D, Schumm T, Metz L (2017) Began: boundary equilibrium generative
  adversarial networks. arXiv preprint arXiv:170310717

\bibitem[{Besl and McKay(1992)}]{besl1992method}
Besl PJ, McKay ND (1992) Method for registration of 3-d shapes. In: Sensor
  Fusion IV: Control Paradigms and Data Structures, vol 1611, pp 586--607

\bibitem[{Booth and Zafeiriou(2014)}]{booth2014optimal}
Booth J, Zafeiriou S (2014) Optimal uv spaces for facial morphable model
  construction. In: Proceedings of the IEEE International Conference on Image
  Processing (ICIP), pp 4672--4676

\bibitem[{Booth et~al.(2016)Booth, Roussos, Zafeiriou, Ponniah, and
  Dunaway}]{booth20163d}
Booth J, Roussos A, Zafeiriou S, Ponniah A, Dunaway D (2016) A 3d morphable
  model learnt from 10,000 faces. In: Proceedings of the IEEE Conference on
  Computer Vision and Pattern Recognition, pp 5543--5552

\bibitem[{Bousmalis et~al.(2017)Bousmalis, Silberman, Dohan, Erhan, and
  Krishnan}]{bousmalis2017unsupervised}
Bousmalis K, Silberman N, Dohan D, Erhan D, Krishnan D (2017) Unsupervised
  pixel-level domain adaptation with generative adversarial networks. In:
  Proceedings of the IEEE Conference on Computer Vision and Pattern Recognition
  (CVPR), vol~1, p~7

\bibitem[{Bronstein et~al.(2017)Bronstein, Bruna, LeCun, Szlam, and
  Vandergheynst}]{bronstein2017geometric}
Bronstein MM, Bruna J, LeCun Y, Szlam A, Vandergheynst P (2017) Geometric deep
  learning: going beyond euclidean data. IEEE Signal Processing Magazine
  34(4):18--42

\bibitem[{Brunton et~al.(2014)Brunton, Salazar, Bolkart, and
  Wuhrer}]{brunton2014review}
Brunton A, Salazar A, Bolkart T, Wuhrer S (2014) Review of statistical shape
  spaces for 3d data with comparative analysis for human faces. Computer Vision
  and Image Understanding 128:1--17

\bibitem[{Cheng et~al.(2018)Cheng, Kotsia, Pantic, and
  Zafeiriou}]{cheng20184dfab}
Cheng S, Kotsia I, Pantic M, Zafeiriou S (2018) 4dfab: A large scale 4d
  database for facial expression analysis and biometric applications. In:
  Proceedings of the IEEE conference on computer vision and pattern recognition
  (CVPR), pp 5117--5126

\bibitem[{Choi et~al.(2017)Choi, Choi, Kim, Ha, Kim, and
  Choo}]{choi2017stargan}
Choi Y, Choi M, Kim M, Ha JW, Kim S, Choo J (2017) Stargan: Unified generative
  adversarial networks for multi-domain image-to-image translation. arXiv
  preprint 1711

\bibitem[{Clevert et~al.(2015)Clevert, Unterthiner, and
  Hochreiter}]{clevert2015fast}
Clevert DA, Unterthiner T, Hochreiter S (2015) Fast and accurate deep network
  learning by exponential linear units (elus). arXiv preprint arXiv:151107289

\bibitem[{Davies et~al.(2008)Davies, Twining, and
  Taylor}]{davies2008statistical}
Davies R, Twining C, Taylor C (2008) Statistical models of shape: Optimization
  and evaluation. Springer Science \& Business Media

\bibitem[{De~Smet and Van~Gool(2010)}]{de2010optimal}
De~Smet M, Van~Gool L (2010) Optimal regions for linear model-based 3d face
  reconstruction. In: Proceedings of the Asian Conference on Computer Vision,
  pp 276--289

\bibitem[{Dosovitskiy and Brox(2016)}]{dosovitskiy2016generating}
Dosovitskiy A, Brox T (2016) Generating images with perceptual similarity
  metrics based on deep networks. In: Proceedings of the Advances in Neural
  Information Processing Systems (NIPS), pp 658--666

\bibitem[{Dou et~al.(2017)Dou, Shah, and Kakadiaris}]{dou2017end}
Dou P, Shah SK, Kakadiaris IA (2017) End-to-end 3d face reconstruction with
  deep neural networks. In: Proceedings of the IEEE Conference on Computer
  Vision and Pattern Recognition (CVPR), pp 21--26

\bibitem[{Fan et~al.(2017)Fan, Su, and Guibas}]{fan2017point}
Fan H, Su H, Guibas LJ (2017) A point set generation network for 3d object
  reconstruction from a single image. In: Proceedings of the IEEE Conference on
  Computer Vision and Pattern Recognition (CVPR), vol~2, p~6

\bibitem[{Feng et~al.(2018)Feng, Wu, Shao, Wang, and Zhou}]{feng2018joint}
Feng Y, Wu F, Shao X, Wang Y, Zhou X (2018) Joint 3d face reconstruction and
  dense alignment with position map regression network. arXiv preprint
  arXiv:180307835

\bibitem[{Genova et~al.(2018)Genova, Cole, Maschinot, Sarna, Vlasic, and
  Freeman}]{genova2018unsupervised}
Genova K, Cole F, Maschinot A, Sarna A, Vlasic D, Freeman WT (2018)
  Unsupervised training for 3d morphable model regression. In: Proceedings of
  the IEEE Conference on Computer Vision and Pattern Recognition (CVPR), pp
  8377--8386

\bibitem[{Goodfellow et~al.(2014)Goodfellow, Pouget-Abadie, Mirza, Xu,
  Warde-Farley, Ozair, Courville, and Bengio}]{goodfellow2014generative}
Goodfellow I, Pouget-Abadie J, Mirza M, Xu B, Warde-Farley D, Ozair S,
  Courville A, Bengio Y (2014) Generative adversarial nets. In: Proceedings of
  the Advances in neural information processing systems, pp 2672--2680

\bibitem[{Gower(1975)}]{gower1975generalized}
Gower JC (1975) Generalized procrustes analysis. Psychometrika 40(1):33--51

\bibitem[{He et~al.(2016)He, Zhang, Ren, and Sun}]{he2016deep}
He K, Zhang X, Ren S, Sun J (2016) Deep residual learning for image
  recognition. In: Proceedings of the IEEE conference on computer vision and
  pattern recognition (CVPR), pp 770--778

\bibitem[{Huang et~al.(2017)Huang, Liu, Van Der~Maaten, and
  Weinberger}]{huang2017densely}
Huang G, Liu Z, Van Der~Maaten L, Weinberger KQ (2017) Densely connected
  convolutional networks. In: Proceedings of the IEEE Conference on Computer
  Vision and Pattern Recognition (CVPR), vol~1, p~3

\bibitem[{Isola et~al.(2017)Isola, Zhu, Zhou, and Efros}]{isola2017image}
Isola P, Zhu JY, Zhou T, Efros AA (2017) Image-to-image translation with
  conditional adversarial networks. arXiv preprint

\bibitem[{Jackson et~al.(2017)Jackson, Bulat, Argyriou, and
  Tzimiropoulos}]{jackson2017large}
Jackson AS, Bulat A, Argyriou V, Tzimiropoulos G (2017) Large pose 3d face
  reconstruction from a single image via direct volumetric cnn regression. In:
  Proceedings of the IEEE International Conference on Computer Vision (ICCV),
  pp 1031--1039

\bibitem[{Johnson et~al.(2016)Johnson, Alahi, and
  Fei-Fei}]{johnson2016perceptual}
Johnson J, Alahi A, Fei-Fei L (2016) Perceptual losses for real-time style
  transfer and super-resolution. In: Proceedings of the European Conference on
  Computer Vision, Springer, pp 694--711

\bibitem[{Jolliffe(2011)}]{jolliffe2011principal}
Jolliffe I (2011) Principal component analysis. In: International Encyclopedia
  of Statistical Science, Springer, pp 1094--1096

\bibitem[{Karras et~al.(2018)Karras, Aila, Laine, and
  Lehtinen}]{karras2017progressive}
Karras T, Aila T, Laine S, Lehtinen J (2018) Progressive growing of gans for
  improved quality, stability, and variation. Proceedings of the International
  Conference for Learning Representations (ICLR)

\bibitem[{Kim et~al.(2017)Kim, Cha, Kim, Lee, and Kim}]{kim2017learning}
Kim T, Cha M, Kim H, Lee JK, Kim J (2017) Learning to discover cross-domain
  relations with generative adversarial networks. arXiv preprint
  arXiv:170305192

\bibitem[{Kingma and Ba(2014)}]{kingma2014adam}
Kingma DP, Ba J (2014) Adam: A method for stochastic optimization. arXiv
  preprint arXiv:14126980

\bibitem[{Kingma and Welling(2013)}]{kingma2013auto}
Kingma DP, Welling M (2013) Auto-encoding variational bayes. arXiv preprint
  arXiv:13126114

\bibitem[{Ledig et~al.(2017)Ledig, Theis, Husz{\'a}r, Caballero, Cunningham,
  Acosta, Aitken, Tejani, Totz, Wang et~al.}]{ledig2017photo}
Ledig C, Theis L, Husz{\'a}r F, Caballero J, Cunningham A, Acosta A, Aitken AP,
  Tejani A, Totz J, Wang Z, et~al. (2017) Photo-realistic single image
  super-resolution using a generative adversarial network. In: Proceedings of
  the IEEE Conference on Computer Vision and Pattern Recognition (CVPR), vol~2,
  p~4

\bibitem[{Lei et~al.(2017)Lei, Jin, Barzilay, and Jaakkola}]{lei2017deriving}
Lei T, Jin W, Barzilay R, Jaakkola T (2017) Deriving neural architectures from
  sequence and graph kernels. arXiv preprint arXiv:170509037

\bibitem[{Li et~al.(2017)Li, Liu, Yang, and Yang}]{li2017generative}
Li Y, Liu S, Yang J, Yang MH (2017) Generative face completion. In: Proceedings
  of the the IEEE Conference on Computer Vision and Pattern Recognition (CVPR),
  vol~1, p~3

\bibitem[{Litany et~al.(2017{\natexlab{a}})Litany, Bronstein, Bronstein, and
  Makadia}]{litany2017deformable}
Litany O, Bronstein A, Bronstein M, Makadia A (2017{\natexlab{a}}) Deformable
  shape completion with graph convolutional autoencoders. arXiv preprint
  arXiv:171200268

\bibitem[{Litany et~al.(2017{\natexlab{b}})Litany, Remez, Rodol{\`a},
  Bronstein, and Bronstein}]{litany2017deep}
Litany O, Remez T, Rodol{\`a} E, Bronstein AM, Bronstein MM
  (2017{\natexlab{b}}) Deep functional maps: Structured prediction for dense
  shape correspondence. In: Proceedings of the IEEE International Conference on
  Computer Vision (ICCV), pp 5660--5668

\bibitem[{Lucic et~al.(2018)Lucic, Kurach, Michalski, Gelly, and
  Bousquet}]{lucic2017gans}
Lucic M, Kurach K, Michalski M, Gelly S, Bousquet O (2018) Are gans created
  equal? a large-scale study. Proceedings of the Advances in Neural Information
  Processing Systems (NIPS)

\bibitem[{Mahendran and Vedaldi(2015)}]{mahendran2015understanding}
Mahendran A, Vedaldi A (2015) Understanding deep image representations by
  inverting them. In: Proceedings of the IEEE Conference on Computer Vision and
  Pattern Recognition (CVPR), pp 5188--5196

\bibitem[{Maron et~al.(2017)Maron, Galun, Aigerman, Trope, Dym, Yumer, Kim, and
  Lipman}]{maron2017convolutional}
Maron H, Galun M, Aigerman N, Trope M, Dym N, Yumer E, Kim VG, Lipman Y (2017)
  Convolutional neural networks on surfaces via seamless toric covers. ACM
  Transactions on Graphics 36(4):71

\bibitem[{Mirza and Osindero(2014)}]{mirza2014conditional}
Mirza M, Osindero S (2014) Conditional generative adversarial nets. arXiv
  preprint arXiv:14111784

\bibitem[{Newcombe et~al.(2011)Newcombe, Izadi, Hilliges, Molyneaux, Kim,
  Davison, Kohi, Shotton, Hodges, and Fitzgibbon}]{newcombe2011kinectfusion}
Newcombe RA, Izadi S, Hilliges O, Molyneaux D, Kim D, Davison AJ, Kohi P,
  Shotton J, Hodges S, Fitzgibbon A (2011) Kinectfusion: Real-time dense
  surface mapping and tracking. In: Proceedings of the IEEE international
  symposium on Mixed and Augmented Reality (ISMAR), pp 127--136

\bibitem[{Nguyen et~al.(2018)Nguyen, Fookes, Sridharan, Tistarelli, and
  Nixon}]{nguyen2018super}
Nguyen K, Fookes C, Sridharan S, Tistarelli M, Nixon M (2018) Super-resolution
  for biometrics: A comprehensive survey. Pattern Recognition 78:23--42

\bibitem[{Qi et~al.(2017)Qi, Su, Mo, and Guibas}]{qi2017pointnet}
Qi CR, Su H, Mo K, Guibas LJ (2017) Pointnet: Deep learning on point sets for
  3d classification and segmentation. Proceedings of the IEEE Conference on
  Computer Vision and Pattern Recognition (CVPR) 1(2):4

\bibitem[{Radford et~al.(2015)Radford, Metz, and
  Chintala}]{radford2015unsupervised}
Radford A, Metz L, Chintala S (2015) Unsupervised representation learning with
  deep convolutional generative adversarial networks. arXiv preprint
  arXiv:151106434

\bibitem[{Ranjan et~al.(2018)Ranjan, Bolkart, Sanyal, and
  Black}]{ranjan2018generating}
Ranjan A, Bolkart T, Sanyal S, Black MJ (2018) Generating 3d faces using
  convolutional mesh autoencoders. arXiv preprint arXiv:180710267

\bibitem[{Richardson et~al.(2017)Richardson, Sela, Or-El, and
  Kimmel}]{richardson2017learning}
Richardson E, Sela M, Or-El R, Kimmel R (2017) Learning detailed face
  reconstruction from a single image. In: Proceedings of the IEEE Conference on
  Computer Vision and Pattern Recognition (CVPR), pp 5553--5562

\bibitem[{Tran et~al.(2017)Tran, Hassner, Masi, and
  Medioni}]{tran2017regressing}
Tran AT, Hassner T, Masi I, Medioni G (2017) Regressing robust and
  discriminative 3d morphable models with a very deep neural network. In:
  Proceedings of the IEEE Conference on Computer Vision and Pattern Recognition
  (CVPR), pp 1493--1502

\bibitem[{Tzeng et~al.(2017)Tzeng, Hoffman, Saenko, and
  Darrell}]{tzeng2017adversarial}
Tzeng E, Hoffman J, Saenko K, Darrell T (2017) Adversarial discriminative
  domain adaptation. In: Proceedings of the IEEE Conference Computer Vision and
  Pattern Recognition (CVPR), vol~1, p~4

\bibitem[{Wang et~al.(2018)Wang, Liu, Zhu, Tao, Kautz, and
  Catanzaro}]{wang2018high}
Wang TC, Liu MY, Zhu JY, Tao A, Kautz J, Catanzaro B (2018) High-resolution
  image synthesis and semantic manipulation with conditional gans. In:
  Proceedings of the IEEE Conference on Computer Vision and Pattern Recognition
  (CVPR), vol~1, p~5

\bibitem[{Wang et~al.(2017)Wang, Huang, You, Yang, and Neumann}]{wang2017shape}
Wang W, Huang Q, You S, Yang C, Neumann U (2017) Shape inpainting using 3d
  generative adversarial network and recurrent convolutional networks. arXiv
  preprint arXiv:171106375

\bibitem[{Yang et~al.(2017)Yang, Lu, Lin, Shechtman, Wang, and
  Li}]{yang2017high}
Yang C, Lu X, Lin Z, Shechtman E, Wang O, Li H (2017) High-resolution image
  inpainting using multi-scale neural patch synthesis. In: Proceedings of the
  IEEE Conference on Computer Vision and Pattern Recognition (CVPR), vol~1, p~3

\bibitem[{Zhao et~al.(2016)Zhao, Mathieu, and LeCun}]{zhao2016energy}
Zhao J, Mathieu M, LeCun Y (2016) Energy-based generative adversarial network.
  arXiv preprint arXiv:160903126

\bibitem[{Zhu et~al.(2017)Zhu, Park, Isola, and Efros}]{zhu2017unpaired}
Zhu JY, Park T, Isola P, Efros AA (2017) Unpaired image-to-image translation
  using cycle-consistent adversarial networks. arXiv preprint

\end{thebibliography}
}
\end{sloppypar}
\end{document}